\documentclass{article}

\usepackage{microtype}
\usepackage{graphicx}
\usepackage{subcaption}
\usepackage{booktabs} 

\usepackage{hyperref}

\usepackage[preprint]{icml2026}

\usepackage{amsmath}
\usepackage{amssymb}
\usepackage{mathtools}
\usepackage{amsthm}

\usepackage[capitalize,noabbrev]{cleveref}

\theoremstyle{plain}

\theoremstyle{definition}

\theoremstyle{remark}

\usepackage{xurl}

\usepackage[textsize=tiny]{todonotes}
\usepackage{amsthm}
\usepackage{tabularx}
\usepackage{threeparttable}
\usepackage{placeins}
\icmltitlerunning{StefaLand: Efficient Geoscience Representation Learning Model for Dynamic Land-Surface Prediction}
\begin{document}

\twocolumn[
\icmltitle{StefaLand: An Efficient Geoscience Foundation Model That Improves Dynamic Land-Surface Predictions}

  \icmlsetsymbol{equal}{*}

    \begin{icmlauthorlist}
      \icmlauthor{Nicholas Kraabel}{equal,psu_eng}
      \icmlauthor{Jiangtao Liu}{equal,psu_eng}
      \icmlauthor{Daniel Kifer}{psu_ces}
      \icmlauthor{Yuchen Bian}{amazon_res}
      \icmlauthor{Chaopeng Shen}{psu_eng}
    \end{icmlauthorlist}
    \icmlaffiliation{psu_eng}{Department of Civil and Environmental Engineering, The Pennsylvania State University, University Park, PA 16802-1408, USA}
    \icmlaffiliation{psu_ces}{Department of Computer Science and Engineering, The Pennsylvania State University, University Park, PA 16802-1408, USA}
    \icmlaffiliation{amazon_res}{Amazon.com, Inc.}
    
  \icmlcorrespondingauthor{Shen Chaopeng}{cxs1024@psu.edu}

\icmlkeywords{Machine Learning, Earth System Science, Hydrology, Land Surface Modeling}
\vskip 0.3in
]
\printAffiliationsAndNotice{\icmlEqualContribution}

\begin{abstract}
Managing natural resources and mitigating risks from floods, droughts, wildfires, and landslides require models that can accurately predict climate-driven land–surface responses. Traditional models often struggle with spatial generalization because they are trained/calibrated on limited observations and can degrade under concept drift. Recently proposed vision foundation models trained on satellite imagery demand massive compute, and they are not designed for dynamic land surface prediction tasks. We introduce StefaLand, a generative spatiotemporal Earth representation learning model centered on learning cross-domain interactions to suppress overfitting. StefaLand demonstrates especially strong spatial generalization on five datasets across four important tasks: streamflow, soil moisture, soil composition and landslides, compared to previous state-of-the-art methods. The domain-inspired design choices include a location-aware masked autoencoder that fuses static and time-series inputs, an attribute-based rather than image-based representation that drastically reduces compute demands, and residual fine-tuning adapters that strengthen knowledge transfer across tasks. StefaLand can be pretrained and finetuned on commonly-available academic compute resources, yet consistently outperforms state-of-the-art supervised learning baselines, fine-tuned vision foundation models and commercially-available embeddings, highlighting the previously overlooked value of cross-domain interactions and providing assistance to data-poor regions of the world. 
\end{abstract}

\section{Introduction}
\label{sec:intro}

Climate change is ushering in strong and widespread changes on the land surface, including higher frequencies of floods, droughts, wildfires and other geohazards \citep{ebi2021extreme, ipcc2021summary}. To mitigate the impact of these disasters, there are urgent needs for models that can accurately predict land surface dynamics such as streamflow, soil moisture, soil composition, landslides, snow water equivalent, groundwater levels, and vegetation carbon content. Among these, soil moisture controls ecosystem health and influences land-atmosphere interactions \citep{dorigo2013global}. Streamflow is the flow rate of water running in the rivers, the most accessible water resource to humans, and too high or too low streamflow can cause flooding or hydrologic drought, respectively. Soil composition (sand, silt, clay fractions) governs infiltration capacity and root-zone storage, while slope–soil-vegetation interactions directly influence landslide hazards. Here, we limit our scope to the predictions of dynamical or static land surface processes that represent the impacts of climate change.

Traditionally, these tasks were undertaken by physics-based models that take atmospheric forcings (precipitation, temperature) as inputs and sequentially calculate the physical processes that eventually lead to the variables of interest \citep{li2015evaluating}. In recent years, there has been a proliferation of data-driven machine learning (ML) models \citep{solomatine2008data}. These models are often set up to accept forcing (dynamic weather) and landscape characteristics (static) data as inputs, and are trained to directly predict the natural land surface variables given the weather inputs. However, up to now, most of the geoscientific ML models have been supervised ML approaches trained specifically for a narrow set of tasks.

Large vision foundation models have been trained on satellite imagery of earth to facilitate mapping \cite{tseng2025galileo}, weather prediction \cite{schmude2024prithviwxc}, air quality \cite{bodnar2025aurora} and other geoscientific tasks \citep{jakubik2023foundation,jakubik2025terramind}. A particularly notable observation is that they have not focused on the interactions among landscape domains (climate, soil, vegetation, terrain, geology). The catchment coevolution hypothesis \cite{Troch2015} states that terrain, soil, vegetation and climate all coevolve, shaping the landscape we have today. Knowing parts of the landscape domains often allow us to predict the others. This implies that their joint distributions can greatly inform latent processes relevant to the transport of water and materials in the catchments. However, valuable temporal datasets and ground-based observations remain underutilized \citep{xie2023geofoundation}, and to our knowledge, no foundation model has yet been developed with a primary focus on dynamical land surface modeling.

A Grand Challenge for geoscientific ML models is to improve their spatial generalization, because a frequent issue facing them is the sparsity and spatial imbalance of observational data. Satellite data are often coarse in resolution and uncertain compared to in-situ measurements. SMAP, for example, provides global soil moisture observations at ~9–36 km resolution every 2–3 days, which are useful for regional climate and hydrologic research but far less valuable than in-situ probes for operational field tasks such as irrigation scheduling or crop stress monitoring \citep{smap2010, liu2022multiscale, liu2023gmd}. However, in-situ data, due to the cost of installing instruments and varying policies on data sharing, is only available in high density in certain regions of some developed nations. For example, streamflow gauge data are abundant in the United States, Europe, Australia and Japan, but remain sparse in Africa, South America, and much of Asia \citep{GRDC2020}. As quantified in many studies~\citep{feng2023suitability}, a deep network trained on data from some regions can face substantial performance degradation when applied in data-scarce regions. This occurs partly because there are not enough sites to learn the true dependencies of the targets on static land surface characteristics, and partly because of systematic data discrepancies across regions (concept drift). While such limitations hinder traditional supervised ML models, foundation or representation learning models offer a potential path forward: by jointly learning from broad, heterogeneous datasets (including temporal records and ground observations where available), they may transfer useful representations to data-scarce regions where task-specific training data are limited.

\textbf{Related Work:} In hydrologic and ecosystem predictions, supervised long short-term memory (LSTM) networks \citep{hochreiter1997lstm} remain a highly popular architecture, in part because land surface processes often behave like Markov processes where LSTMs’ gating mechanisms handle noisy continuous inputs well \citep{Kratzert2018HESS}. Attempts to adopt transformers, so successful in natural language processing, have generally found it difficult to noticeably surpass LSTM in time series regression tasks \citep{xue2023make,liu2024jhydrol, liu2025rnns}, with evidence of overfitting on continuous signals \citep{zeng2022transformers}. Nonetheless, recent studies show that with task-specific modifications and careful fine-tuning, transformers can achieve competitive results in extreme event prediction \citep{wen2023transformers}, precisely the areas where current hydrologic models struggle most with.

Traditional hydrologic research on "prediction in ungauged basins" (PUB) have examined regionalization and spatial interpolation approaches including clustering or classifying catchments and transferring parameters from donor catchments in the same class \citep{hrachowitz2013pub,yang2023regionalization}. Such an expert-derived design represents a crude practice of unsupervised learning that indicates the importance of understanding the joint data distribution. However, modern weakly-supervised foundation models can, in general, much better grasp the joint data distribution than expert-driven approaches. Representation learning models offer a promising approach to address these spatial generalization challenges. By pretraining on large-scale datasets to learn generalizable representations, these models can potentially transfer knowledge across regions and geoscientific domains \citep{zhang2024when}. 

Recent progress in geoscientific foundation models has been driven by the increasing availability of global Earth system datasets and advances in self-supervised learning, enabling pretrained representations that transfer across tasks and regions \citep{bommasani2022oundationmodels,lacoste2023geobench}. Beyond vision-based approaches, recent work on structured data has shown that foundation models such as TabPFN can learn transferable priors over tabular regression tasks without task-specific training. \citep{hollmann2025tabular}.

Most existing Earth observation foundation models, including TerraMind \citep{jakubik2025terramind}, Prithvi \citep{hsu2024geospatial}, Aurora \citep{bodnar2025aurora}, AlphaEarth \citep{brown2025alphaearth}, and Galileo \citep{tseng2025galileo}, are pretrained primarily on remote sensing imagery and related EO products. These models are effective at capturing surface appearance and spatial patterns, and their representations often act as static or slowly varying landscape descriptors. In particular, AlphaEarth learns global, high-resolution spatial embeddings from large-scale Earth observation data that capture consistent landscape structure across climate and biome gradients, and these embeddings have been shown to transfer effectively as fixed spatial features for a wide range of downstream Earth system prediction tasks, including hydrology. Because AlphaEarth embeddings are globally available and independent of task-specific labels, they provide a strong and widely applicable baseline for evaluating spatial generalization in data-limited settings. However, many variables central to land-surface and hydrologic dynamics, including subsurface properties, storage processes, and long-term temporal interactions, are not directly observable from space and therefore cannot be reliably inferred from imagery alone. Moreover, EO foundation models are typically optimized for surface-level semantic consistency and invariance to transient atmospheric or observational effects, which can further suppress signals that are informative for hydrologic processes, such as moisture-related variability or persistence. 

\textbf{Our contributions:} We present the Spatial-Temporal Earth Foundation model with Attributes for the Land Surface (StefaLand), a land-focused geoscientific representation learning model for dynamic land–surface prediction. StefaLand is evaluated on streamflow, soil moisture, soil composition, and landslide susceptibility under spatial holdout regimes. It shows strong spatial generalization across diverse landscapes and data-scarce regions for a wide variety of tasks. StefaLand's attribute-based rather than image-based design (with the potential to link to image-like inputs in the future) incorporates a variety of ground-based measurement data, emphasizes relevant land-surface physical processes, drastically reduces compute requirements while retaining global coverage, making it accessible to researchers with modest resources. Pretraining our model required only about 720 V100 GPU hours (could be shorter with more advanced GPUs). The model builds on a masked autoencoder backbone, a location-aware fusion of static and time-series inputs, grouped masking to promote cross-domain interactions, and residual fine-tuning adapters, into a coherent design guided by geoscientific knowledge. Taken together, these contributions establish StefaLand as an efficient and accessible complement to vision-based foundation models.

\begin{figure*}[t]
    \centering
    \includegraphics[width=\textwidth]{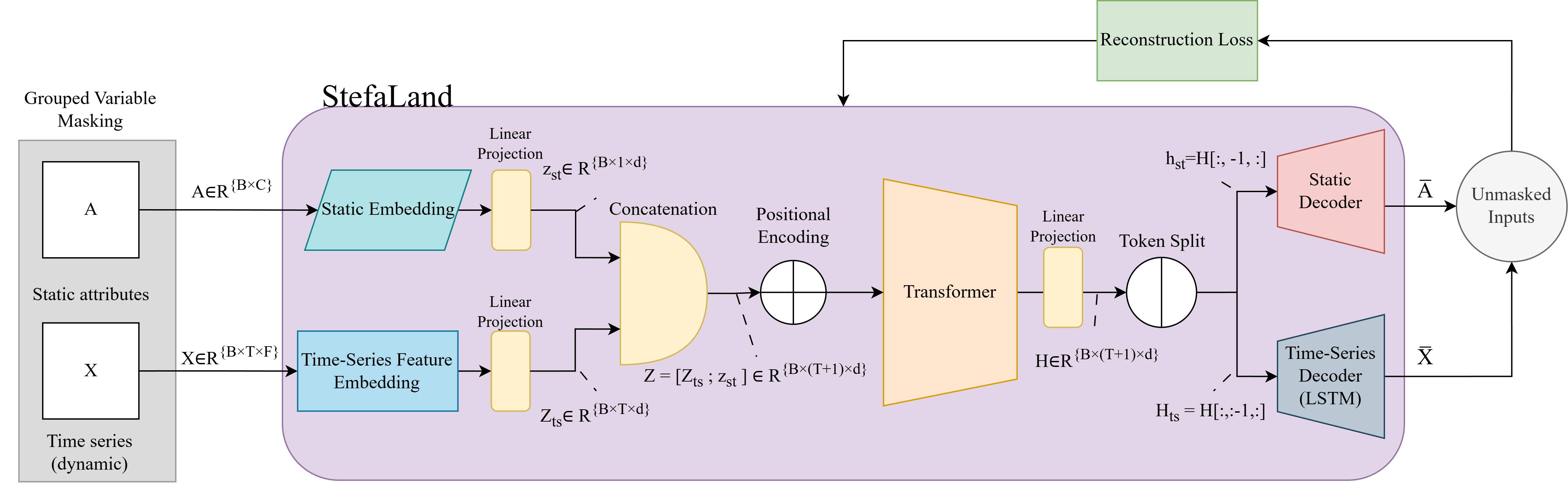}
    \caption{
    Conceptual overview of the StefaLand Structure.
    Static landscape attributes and dynamic forcings are jointly embedded using a transformer-based masked autoencoder with cross-variable group masking. With relevant dimensionality included. 
    }
    \label{fig:StefaLand_overview}
\end{figure*}

\section{Methods} 
\label{sec:methods} 

Dynamic land–surface prediction requires combining heterogeneous information: static landscape attributes such as topography, soils, vegetation, and geology, together with dynamic forcings such as precipitation and temperature. StefaLand addresses this challenge with a transformer-based masked autoencoder that jointly embeds static and dynamic variables, pretrained with a cross-variable masking strategy, and then adapted for specific prediction tasks with task-specific heads.

\subsection{Stefaland Structure}

StefaLand is a transformer-based masked autoencoder inspired by bidirectional language models such as BERT \cite{BERT2019}, designed to jointly embed static landscape attributes and dynamic time-varying forcings within a unified representation. During pretraining, StefaLand randomly masks subsets of the input variables and learns to reconstruct the masked components using the remaining unmasked information. Masking may affect static attributes, temporal variables, or entire variable groups, encouraging the model to learn the joint distribution over heterogeneous land-surface controls rather than relying on any single data source.

A key design choice in StefaLand is cross-variable group masking, in which physically or statistically related variables are masked together rather than independently. This prevents trivial reconstruction through correlated inputs and encourages the model to capture cross-domain linkages, such as interactions between soil texture and climate seasonality or between topography and hydrologic response. Grouped masking strategies have previously been explored in multimodal learning settings, including Presto \cite{tseng2023lightweight}, and here we adapt this idea using a reconstruction loss applied to masked inputs, normalized by variable-wise standard deviations when available. The attention mechanism naturally accommodates missing inputs by suppressing attention weights for masked tokens, allowing StefaLand to flexibly reason over arbitrarily incomplete input configurations. Although more recent masked-model formulations exist, we adopt a BERT-style design for its simplicity, interpretability, and robustness across heterogeneous inputs.

Following pretraining, the encoder produces contextualized embeddings for both static attributes and dynamic forcings, which are reused across downstream tasks through lightweight task-specific heads and residual adaptation pathways, as illustrated in Figure~\ref{fig:StefaLand_overview}.

\subsection{Pretraining Details}

The pretraining dataset is a derived global attribute dataset spanning $\sim$8{,}634 locations (basins) over 40 years. Variables were chosen to represent the key controls on fluxes of water, energy, momentum, sediment, and nutrients. A complete list of variables, their group assignments, and their sources is provided in Appendix ~\ref{appendix:C}. A visualization of the pretraining dataset coverage is provided in Appendix \ref{appendix:C} Figure ~\ref{fig:GLOBAL_data_distribution}.  

The cross-variable group masking (CVGM) scheme is designed so that variables with reciprocal or bidirectional relationships are masked together, preventing them from acting as direct predictors for one another. Most groupings are straightforward and common sense, such as masking silt and clay fractions jointly, while a few variables require domain-specific treatment; for example, soil depth is grouped with terrain attributes due to its strong dependence on topographic derivations. The complete set of variables and their group assignments is provided in Table ~\ref{tab:group_masking}. By masking and reconstructing variables at the group level, the model is encouraged to capture cross-domain interactions, such as couplings between soil texture and climate seasonality.

\subsection{Finetuning for Prediction Tasks}

We adapt the pretrained StefaLand encoder to downstream prediction tasks using lightweight task-specific heads while keeping the main model weights frozen. This strategy, illustrated in Figure~\ref{fig:StefaLand_overview}, preserves the general-purpose representations learned during pretraining and reduces the risk of overfitting when finetuning on limited task-specific data.

Our primary finetuning configuration, \textbf{StefaLand-resConn}, integrates pretrained embeddings with raw meteorological forcings through a residual adaptation pathway via additive fusion prior to the recurrent decoder. Let $E_t$ denote the StefaLand embedding at time $t$, and let $x_t$ denote the corresponding raw forcing inputs. We compute:
\begin{align}
r_t &= f_{\text{conv+linear}}(x_t), \\
h_t &= \text{LSTM}(E_t + r_t), \\
\hat{y}_t &= W_o h_t + b_o ,
\end{align}
where $f_{\text{conv+linear}}(\cdot)$ denotes a shallow convolutional and linear projection block. Residual connections propagate both the pretrained embedding $E_t$ and the task-specific signal $r_t$ into the recurrent decoder, allowing general spatial knowledge captured during pretraining to be iteratively refined using task-specific temporal information. This design strengthens spatial generalization while retaining flexibility to adapt to local dynamics.

We employ an LSTM decoder for tasks with explicit temporal structure, such as streamflow and soil moisture prediction. For non-temporal or spatially structured tasks, including soil property inference and landslide susceptibility mapping, the temporal decoder is replaced with a task-appropriate adapter head, such as a multilayer perceptron or two-dimensional convolutional neural network (CNN2D). In these cases, frozen StefaLand embeddings are used as contextual features. Across all tasks, only the adapter modules and task heads are updated during finetuning, while the pretrained StefaLand encoder remains frozen, ensuring computational efficiency and stable transfer from the representation learning model.


\FloatBarrier

\section{Experiments}
\label{sec:Experimental Design and Results}

We tested the value of foundation model pretraining on 5 datasets and 6 experiments, including, streamflow on the CAMELS dataset on USA (a widely used benchmark dataset in hydrology), CAMELS streamflow prediction with hybrid model, global streamflow on Caravan, global in-situ soil moisture, global soil properties, and landslide susceptibility in Oregon, USA. For all experiments, hyperparameters were tuned with Ray Tune and kept consistent across model configurations within each experimental case (e.g., CAMELS streamflow, soil moisture, etc). Because we compare spatial generalization, we used temporal validation splits for hyperparameter optimization. Complete details of hyperparameters, forcings, and static features for all four experiments are in Appendix~\ref{appendix:C}.

\paragraph{Model variants and baselines.}
We evaluate StefaLand alongside established baselines commonly used in hydrologic and geophysical time series modeling. As supervised baselines, we include LSTM-SL, which remains the dominant architecture for streamflow prediction and follows the same core model class used in prior large scale hydrology studies \citep{kratzert2019learning,feng2021mitigating,sabzipour2023}. We additionally evaluate Informer \citep{zhou2021informer}, Reformer \citep{kitaev2020reformer}, and DLinear \citep{zeng2022transformers}, which, although originally proposed for forecasting, are applied here in a sequence to sequence setting and are commonly included in recent hydrologic and geophysical modeling studies as representative transformer-based baselines.

To assess pretrained Earth representations, we include AlphaEarth LSTM and AlphaEarth ResConn, which feed pretrained AlphaEarth embeddings into an LSTM decoder with and without a residual task adapter, respectively \citep{brown2025alphaearth}. Because our task periods do not fully overlap AlphaEarth’s 2017–2024 temporal coverage, AlphaEarth embeddings are spatially aggregated over each basin and collapsed into a fixed set of 64 static features per site. For CAMELS streamflow and soil moisture, we additionally evaluate TabPFN, a tabular foundation model \citep{hollmann2025tabular}. Since TabPFN is non sequential and constrained by a fixed context window, we provide it with compact statistical summaries of each historical window while preserving identical data splits and evaluation protocols.

We also conducted an exploratory evaluation of additional Earth observation and atmospheric foundation models to assess their viability for land–surface prediction under matched adaptation strategies; these limited-scope results are reported in \ref{appendix:B:foundation}

Our proposed method, StefaLand resConn (residual connection), combines a pretrained encoder with a residual adapter and LSTM decoder.

\subsection{CAMELS Streamflow Prediction}

To compare spatial generalization on a well benchmarked dataset, we follow \citep{feng2021mitigating}, testing prediction in ungauged basins (PUB, random spatial K-fold) and ungauged regions (PUR, regional spatial K-fold). We use CAMELS \citep{addor2017camels,newman2014}, restricted to the 531-basin subset with clear watershed boundaries \citep{newman2017}. Basins were divided into 10 random groups for PUB and 7 contiguous regions for PUR, employing leave-one-out in both cases. To avoid leakage, all CAMELS-overlapping stations were removed during pretraining for PUB, and entire regions were excluded for PUR. The success at these tasks would mean better flood forecasting information for populations in the world who do not have as many gauging stations around them.
\FloatBarrier
\begin{table}[t]
\centering
\caption{CAMELS streamflow performance under PUB evaluation (random spatial holdout; ungauged basins). Values are median $\pm$ standard error across folds.}
\label{tab:camels_pub}
\resizebox{\columnwidth}{!}{%
\begin{tabular}{lcccc}
\toprule
Model & RMSE $\downarrow$ & ubRMSE $\downarrow$ & Corr $\uparrow$ & NSE $\uparrow$ \\
\midrule
LSTM SL  & 1.402 $\pm$ 0.04 & 1.360 $\pm$ 0.04 & 0.762 $\pm$ 0.01 & 0.636 $\pm$ 0.04 \\
DLinear & 2.012 $\pm$ 0.06 & 2.000 $\pm$ 0.06 & 0.598 $\pm$ 0.01 & 0.302 $\pm$ 0.48 \\
Informer & 2.262 $\pm$ 0.07 & 2.237 $\pm$ 0.08 & 0.521 $\pm$ 0.01 & 0.104 $\pm$ 0.08 \\
Reformer & 1.908 $\pm$ 0.09 & 1.871 $\pm$ 0.09 & 0.718 $\pm$ 0.00 & 0.270 $\pm$ 0.18 \\
TabPFN  & 2.727 $\pm$ 0.09 & 2.725 $\pm$ 0.09 & 0.718 $\pm$ 0.01 & 0.509 $\pm$ 0.014 \\
AlphaEarth LSTM  & 1.409 $\pm$ 0.10 & 1.393 $\pm$ 0.10 & 0.837 $\pm$ 0.11 & 0.618 $\pm$ 0.05 \\
AlphaEarth resConn  
& \underline{1.361 $\pm$ 0.09} & \underline{1.345 $\pm$ 0.09} & \underline{0.859 $\pm$ 0.01} & \underline{0.647 $\pm$ 0.04} \\
\textbf{StefaLand resConn} 
& \textbf{1.111 $\pm$ 0.04} & \textbf{1.068 $\pm$ 0.04} & \textbf{0.869 $\pm$ 0.01} & \textbf{0.717 $\pm$ 0.16} \\
\bottomrule
\end{tabular}
}
\end{table}

\begin{table}[t]
\centering
\caption{CAMELS streamflow performance (PUR: regional spatial holdout; ungauged regions). Values are median $\pm$ standard error across folds.}
\label{tab:camels_streamflow_pur}
\resizebox{\columnwidth}{!}{%
\begin{tabular}{lcccc}
\toprule
Model & RMSE $\downarrow$ & ubRMSE $\downarrow$ & Corr $\uparrow$ & NSE $\uparrow$ \\
\midrule
LSTM SL & \underline{1.609 $\pm$ 0.24} & \underline{1.457 $\pm$ 0.22} & 0.743 $\pm$ 0.02 & \underline{0.554 $\pm$ 0.13} \\
DLinear & 2.019 $\pm$ 0.35 & 1.983 $\pm$ 0.34 & 0.597 $\pm$ 0.03 & 0.290 $\pm$ 0.99 \\
Informer & 2.332 $\pm$ 0.41 & 2.295 $\pm$ 0.37 & 0.497 $\pm$ 0.01 & 0.046 $\pm$ 0.27 \\
Reformer & 2.074 $\pm$ 0.33 & 1.978 $\pm$ 0.31 & 0.686 $\pm$ 0.02 & 0.257 $\pm$ 1.17 \\
TabPFN & 2.709 $\pm$ 1.08 & 2.501 $\pm$ 0.89 & 0.576 $\pm$ 0.11 & 0.328 $\pm$ 0.12 \\
AlphaEarth LSTM  & 1.724 $\pm$ 0.85 & 1.685 $\pm$ 0.82 & 0.780 $\pm$ 0.07 & 0.456 $\pm$ 0.15 \\
AlphaEarth resConn  
& 1.727 $\pm$ 0.74 & 1.684 $\pm$ 0.71 & \underline{0.790 $\pm$ 0.06} & 0.520 $\pm$ 0.10 \\
\textbf{StefaLand resConn} 
& \textbf{1.344 $\pm$ 0.21} & \textbf{1.334 $\pm$ 0.19} & \textbf{0.801 $\pm$ 0.02} & \textbf{0.635 $\pm$ 0.25} \\
\bottomrule
\end{tabular}
}
\end{table}

The results in Tables \ref{tab:camels_pub} and \ref{tab:camels_streamflow_pur} demonstrate that foundation-model pretraining substantially improves spatial generalization relative to purely supervised approaches under both PUB and PUR evaluation. StefaLand-resConn achieves the strongest performance across all reported metrics, reducing RMSE by approximately 20\% relative to the supervised LSTM baseline under PUB and by about 16 to 17\% under PUR, while also yielding consistently higher correlation and NSE. Among the baselines, the supervised LSTM remains competitive and outperforms linear and transformer-based sequence models, while TabPFN performs poorly with substantially higher RMSE and lower NSE despite moderate correlation, indicating that a tabular formulation fails to capture key temporal structure in rainfall runoff dynamics. AlphaEarth-based variants provide notable improvements over the supervised LSTM, particularly in correlation, but remain well below StefaLand-resConn across all metrics, even with residual feature reuse.

We ran additional experiments that hybridize StefaLand with the HBV1.1 physics backbone on the same PUB/PUR splits, testing its ability to parameterize physics-based models. These hybrids achieved up to a 13\% RMSE reduction and a 10\% correlation gain compared to the LSTM–HBV1.1 baseline, showing StefaLand's . By constraining predictions with physics while leveraging StefaLand features, these hybrids further improve upon the general results above and highlight the versatility of the approach. Full results are provided in Appendix~\ref{appendix:B}, Table~\ref{tab:camels_streamflow_physics}. 
 
On a related note, the supervised LSTM is not an easy benchmark to surpass. The original multi-basin LSTM and subsequent large-scale comparisons \citep{kratzert2019learning, feng2021mitigating} showed that vanilla Transformers generally failed to outperform LSTMs on rainfall–runoff prediction and modified transforms essentially tied LSTM \citep[][Table~1 therein]{liu2024jhydrol}. The LSTM NSE values reported here are very similar to those in the domain literature \citep{feng2021mitigating}. Google’s global flood-forecasting system adopts an encoder–decoder LSTM backbone \citep{nearing2024nature}. The gains demonstrated are also substantial: to provide some context, when LSTM raised NSE from 0.64 to 0.73 (without ensemble), it was considered a generational change in predictive performance \cite{Nearing2021, feng2021mitigating}, and no method other than ensembling more models cleanly surpassed default LSTM on the PUB test. 


\FloatBarrier
\subsection{Caravan Global Streamflow}

We designed a global-scale runoff prediction experiment to assess spatial robustness and generalization across diverse hydroclimatic regimes. We use the open-source Caravan dataset \citep{kratzert2023caravan}, a global community dataset that integrates meteorological forcings, catchment attributes, and observed streamflow for river basins worldwide. From the full Caravan archive (16,300 basins), we restricted our evaluation to the official CAMELS-family datasets, yielding 3,278 basins with relatively fuller streamflow records. From this set, we constructed a curated subset of 3,026 basins by excluding those with more than 75\% missing streamflow observations during the 10-year evaluation period from 2010 to 2020, ensuring sufficient data quality for model training and testing. This filtering removes sparsely observed basins while preserving substantial geographic, climatic, and physiographic diversity. We employed a five-fold random spatial holdout protocol to evaluate generalization to unseen basins at the global scale.

\begin{table}[t]
\centering
\caption{Caravan streamflow performance under random spatial holdout. Values are median $\pm$ standard error across folds}
\label{tab:caravan_streamflow_pub}
\resizebox{\columnwidth}{!}{%
\begin{tabular}{lcccc}
\toprule
Model & RMSE $\downarrow$ & ubRMSE $\downarrow$ & Corr $\uparrow$ & NSE $\uparrow$ \\
\midrule
LSTM SL & 1.623 $\pm$ 0.87 & 1.589 $\pm$ 0.84 & 0.496 $\pm$ 0.08 & 0.398 $\pm$ 0.74 \\
DLinear & 1.605 $\pm$ 0.99 & 1.453 $\pm$ 0.99 & 0.512 $\pm$ 0.07 & 0.412 $\pm$ 0.81 \\
Informer & 1.562 $\pm$ 1.04 & 1.471 $\pm$ 0.98 & 0.435 $\pm$ 0.07 & 0.256 $\pm$ 0.92 \\
Reformer & 1.574 $\pm$ 1.06 & 1.507 $\pm$ 0.97 & 0.451 $\pm$ 0.08 & 0.378 $\pm$ 0.87 \\
AlphaEarth LSTM & 1.533 $\pm$ 0.97 & 1.495 $\pm$ 0.89 & 0.523 $\pm$ 0.09 & 0.489 $\pm$ 0.89 \\
AlphaEarth resConn & \textbf{1.424 $\pm$ 1.03} & \underline{1.406 $\pm$ 1.00} & \underline{0.541 $\pm$ 0.07} & \underline{0.514 $\pm$ 0.79} \\
StefaLand resConn & \underline{1.457 $\pm$ 0.87} & \textbf{1.401 $\pm$ 0.83} & \textbf{0.589 $\pm$ 0.06} & \textbf{0.533 $\pm$ 0.70} \\
\bottomrule
\end{tabular}
}
\vspace{-.20 in}
\end{table}

Across the global Caravan benchmark (Table~\ref{tab:caravan_streamflow_pub}), StefaLand-resConn achieves the highest correlation and NSE among all evaluated models, indicating improved agreement with observed runoff dynamics under substantial spatial heterogeneity. While error-based metrics such as RMSE and ubRMSE vary considerably across basins, StefaLand-resConn consistently improves correlation relative to both supervised sequence models and alternative pretrained representations. AlphaEarth-based variants yield noticeable gains over purely supervised baselines, but remain below StefaLand-resConn across most metrics. The Caravan is a more noisy dataset due to global inconsistencies in data collection. These results suggest that attribute-centric pretraining combined with residual temporal adaptation provides a foundation to improve global hydrologic services.

\FloatBarrier
\subsection{Global Soil Moisture}

We evaluated finetuning StefaLand for soil moisture predictions following \cite{liu2023evaluating}, using ISMN \citep{dorigo_international_2011,dorigo2013global}. Even though there is a globally covering satellite-based product for soil moisture, the data quality can hardly match that of in-situ moisture sensors; thus the ability to generalize in-situ data is valuable. ISMN consists of 1,316 ground-based stations. We performed five-fold spatial cross-validation for random holdout and a regional holdout on Europe, training on all other continents while excluding European sites (129) for testing.  LSTM again serves as the established state-of-the-art baseline \citep{wang2024hess,liu2023gmd}.
\begin{table}[t]
\centering
\caption{Soil moisture prediction across ISMN (random location holdout). Values are median $\pm$ standard error across folds.}
\label{tab:ismn_soilmoist_random}
\resizebox{\columnwidth}{!}{%
\begin{tabular}{lccc}
\toprule
Model & RMSE $\downarrow$ & ubRMSE $\downarrow$ & Corr $\uparrow$ \\
\midrule
LSTM SL & \underline{0.073 $\pm$ 0.002} & \underline{0.055 $\pm$ 0.001} & \underline{0.764 $\pm$ 0.006} \\
DLinear & 0.088 $\pm$ 0.001 & 0.065 $\pm$ 0.001 & 0.612 $\pm$ 0.007 \\
Informer & 0.102 $\pm$ 0.002 & 0.082 $\pm$ 0.001 & 0.232 $\pm$ 0.012 \\
Reformer & 0.090 $\pm$ 0.002 & 0.071 $\pm$ 0.001 & 0.568 $\pm$ 0.015 \\
TabPFN & \textbf{0.068 $\pm$ 0.004} & 0.057 $\pm$ 0.003 & 0.404 $\pm$ 0.016 \\
AlphaEarth LSTM & 0.075 $\pm$ 0.001 & 0.062 $\pm$ 0.001 & 0.427 $\pm$ 0.019 \\
AlphaEarth resConn & 0.082 $\pm$ 0.001 & 0.067 $\pm$ 0.001 & 0.406 $\pm$ 0.012 \\
\textbf{StefaLand resConn} & \textbf{0.068 $\pm$ 0.001} & \textbf{0.054 $\pm$ 0.001} & \textbf{0.783 $\pm$ 0.005} \\
\bottomrule
\end{tabular}
}
\end{table}

\begin{table}[t]
\centering
\caption{Soil moisture prediction across ISMN cross-continental validation on Europe (129 sites).}
\label{tab:ismn_soilmoist_europe}
\resizebox{\columnwidth}{!}{%
\begin{tabular}{lccc}
\toprule
Model & RMSE $\downarrow$ & ubRMSE $\downarrow$ & Corr $\uparrow$ \\
\midrule
LSTM SL & 0.112 & \underline{0.053} & 0.510 \\
DLinear & 0.093 & \textbf{0.051} & \underline{0.623} \\
Informer & 0.088 & 0.063 & 0.358 \\
Reformer & \underline{0.087} & 0.058 & 0.553 \\
TabPFN & \textbf{0.081} & 0.054 & 0.401 \\
AlphaEarth LSTM & 0.087 & 0.068 & 0.313 \\
AlphaEarth resConn & 0.090 & 0.071 & 0.308 \\
\textbf{StefaLand resConn} & 0.090 & 0.059 & \textbf{0.638} \\
\bottomrule
\end{tabular}
}
\vspace{-.15 in}
\end{table}

Against these baselines, the soil moisture experiments demonstrate advantages of the StefaLand-resConn architecture (Tables \ref{tab:ismn_soilmoist_random} and \ref{tab:ismn_soilmoist_europe}). For random spatial holdout, StefaLand-resConn attains the highest correlation (0.783) while also achieving the lowest ubRMSE (0.054) and matching the best RMSE (0.068), outperforming the LSTM baseline across all metrics. Although TabPFN achieves comparable RMSE, its substantially lower correlation indicates weaker agreement with temporal variability relative to sequence-based models.

The regional holdout on Europe represents a more challenging cross-continental generalization setting. In this case, StefaLand-resConn again achieves the highest correlation (0.638), exceeding all baselines including LSTM and linear or transformer-based models, while maintaining competitive RMSE and ubRMSE. In contrast, AlphaEarth-based variants show limited transfer performance, with both LSTM and residual-connection formulations yielding lower correlations despite similar error magnitudes.

Overall, these results highlight that residual adaptation on top of geoscience-pretrained representations is critical for robust soil moisture prediction, particularly under strong spatial distribution shift.

\FloatBarrier
\subsection{Soil Property Prediction}

There are different soil datasets, each collected with different protocols and data processing techniques, resulting in significant discrepancies. In this test, we finetuned StefaLand to predict in situ soil profile data from another dataset (ISRIC). This application can produce a seamless dataset that is consistent with a set of in situ data, improving data availability and addressing systematic biases. In addition, it helps us understand the noise associated with each dataset. StefaLand’s pretraining soils dataset is HWSD, which has some overlap but also extensive differences from ISRIC, which is larger and potentially noisier. We finetuned StefaLand to predict one soil texture property (e.g., clay percentage) in ISRIC while masking the corresponding complementary attribute (e.g., sand) from the same profile to avoid information leakage, probing how easy it is to infer soil properties using other attributes such as climate, terrain, and land cover. We compared StefaLand finetuning against AlphaEarth features incorporated into a random forest model, as well as supervised random forest and linear regression baselines. Train test splits can be found here \ref{appendix:dataset_splitting}. 

\begin{table}[t]
\centering
\caption{Soil composition prediction (clay). Values are median $\pm$ standard error across folds.}
\label{tab:soil_comp_clay}
\resizebox{\columnwidth}{!}{%
\begin{tabular}{lcc}
\toprule
Model & Corr $\uparrow$ & $R^2$ $\uparrow$ \\
\midrule
Random Forest & 0.456 $\pm$ 0.02 & \underline{0.207 $\pm$ 0.01} \\
Linear Regression & 0.138 $\pm$ 0.01 & 0.019 $\pm$ 0.01 \\
AlphaEarth RF & \underline{0.462 $\pm$ 0.03} & 0.205 $\pm$ 0.02 \\
\textbf{StefaLand finetuned} & \textbf{0.509 $\pm$ 0.01} & \textbf{0.259 $\pm$ 0.01 \rule{0pt}{2.6ex}} \\
\bottomrule
\end{tabular}
}
\end{table}

\begin{table}[t]
\centering
\caption{Soil composition prediction (sand). Values are median $\pm$ standard error across folds.}
\label{tab:soil_comp_sand}
\resizebox{\columnwidth}{!}{%
\begin{tabular}{lcc}
\toprule
Model & Corr $\uparrow$ & $R^2$ $\uparrow$ \\
\midrule
Random Forest & 0.585 $\pm$ 0.01 & 0.342 $\pm$ 0.01 \\
Linear Regression & 0.347 $\pm$ 0.01 & 0.120 $\pm$ 0.01 \\
AlphaEarth RF & \underline{0.615 $\pm$ 0.01} & \underline{0.373 $\pm$ 0.01} \\
\textbf{StefaLand finetuned} & \textbf{0.704 $\pm$ 0.01} & \textbf{0.495 $\pm$ 0.01\rule{0pt}{2.6ex}} \\
\bottomrule
\end{tabular}
}
\vspace{-.15 in}
\end{table}

StefaLand finetuning achieves the strongest performance for both clay and sand prediction, outperforming linear regression, random forest, and AlphaEarth RF baselines in terms of both correlation and $R^2$. While AlphaEarth RF improves over standard random forest baselines, it remains consistently below StefaLand finetuned across both soil properties. These results indicate that StefaLand’s pretrained representations are particularly effective when adapted to infer soil texture attributes from complementary variables and static environmental context.

\subsection{Landslide Susceptibility Prediction}

Landslide is a geohazard that kill thousands each year. We next evaluated StefaLand for landslide susceptibility prediction using the SLIDO dataset from the State of Oregon, which provides detailed landslide occurrence records. Following \cite{shen2025slido}, this is a binary classification task indicating the presence or absence of landslides in a 30m by 30m patch. 

\begin{table}[h]
\centering
\caption{Landslide susceptibility prediction results on the Oregon SLIDO dataset.}
\label{tab:landslides}
\resizebox{\linewidth}{!}{%
\begin{tabular}{l|ccccc}
\toprule
Model & Accuracy  $\uparrow$ & Precision  $\uparrow$ & Recall  $\uparrow$ & F1  $\uparrow$ & ROC AUC  $\uparrow$ \\
\midrule
Logistic Regression 2D 
& 0.744 & 0.720 & 0.795 & 0.756 & 0.823 \\
Random Forest 2D 
& 0.765 & 0.737 & 0.822 & 0.777 & 0.849 \\
CNN2D 
& \underline{0.880} & \textbf{0.896} & \underline{0.858} & \underline{0.877} & \textbf{0.954} \\
StefaLand + CNN2D
& \textbf{0.903} & \underline{0.859} & \textbf{0.963} & \textbf{0.908} & \underline{0.911} \\
\bottomrule
\end{tabular}
}

\smallskip
\noindent\scriptsize\textit{Note}: All baseline results (Logistic Regression 2D, Random Forest 2D, CNN2D) are taken from previously published 10m-resolution experiments in \cite{shen2025slido}, except for StefaLand + CNN2D, which represents our proposed method.
\vspace{-.15 in}
\end{table}
We finetuned StefaLand by extracting frozen hidden features and concatenating them with a 2D CNN, then retrained the CNN classifier to assess StefaLand’s ability to provide generalizable geoscience features.

Results show that StefaLand’s pretrained features improved the CNN’s generalization, yielding modest gains across most metrics. This is a particularly difficult baseline to improve, so even modest gains are rare. Recall increased from 0.858 to 0.963, and accuracy rose from 0.880 to 0.903, reflecting fewer misclassifications overall. While the random forest achieves high precision in this particular run, it does so at the cost of lower recall. Overall, StefaLand produces well-rounded predictions, with both higher accuracy and recall than CNN2D.

\FloatBarrier
\subsection{Ablations}

\begin{figure}[t]
  \centering
  \includegraphics[width=\columnwidth]{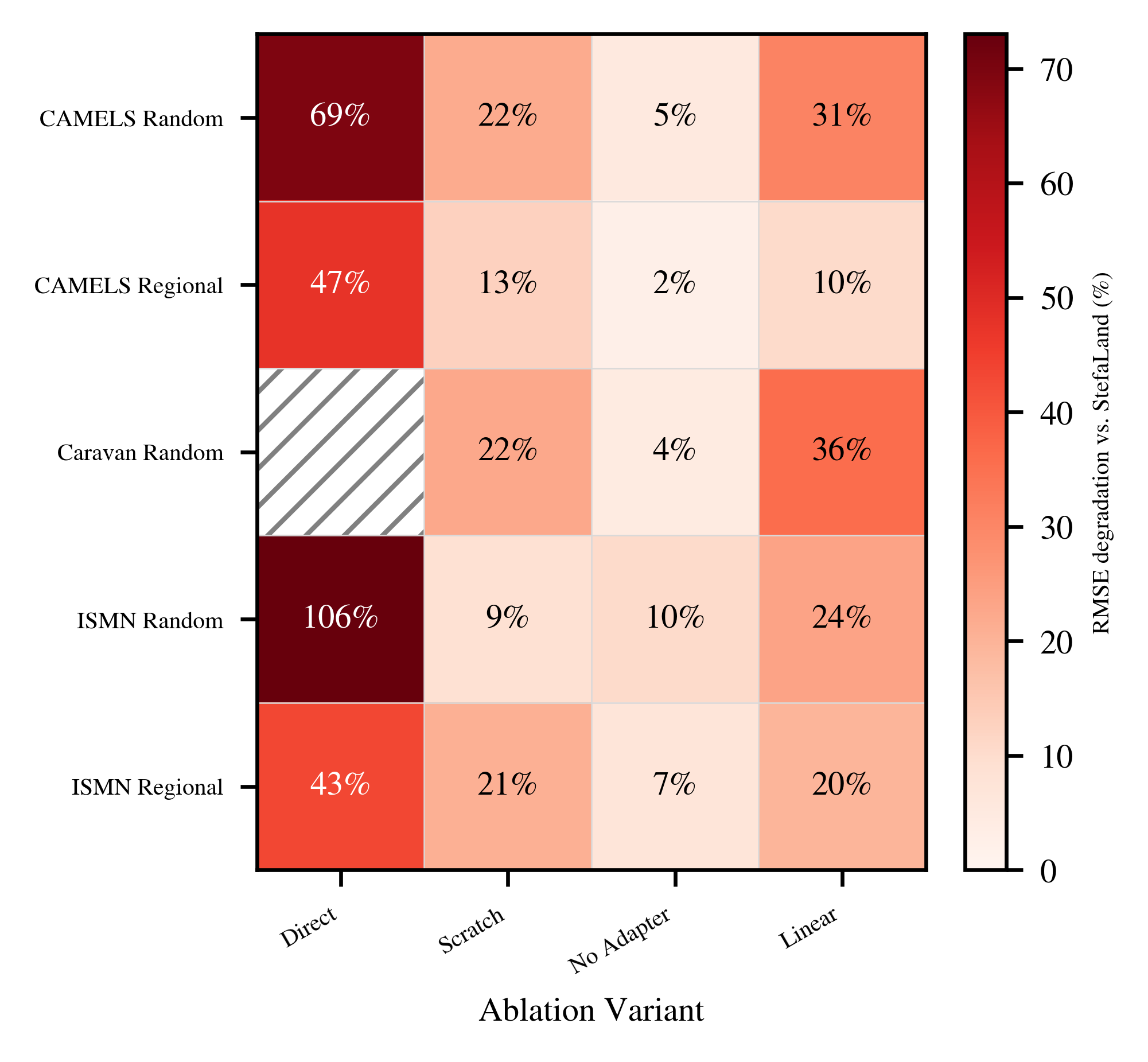}
  \caption{Ablation impact matrix across evaluation settings using RMSE. Each cell shows the percent RMSE increase relative to the full StefaLand ResConn model (lower is better), highlighting the contributions of pretraining and task adaptation. Blank or hatched cells indicate ablations not evaluated for that setting.}
  \label{fig:ablation_rmse_heatmap}
  \vspace{-.2 in}
\end{figure}

To isolate the contributions of pretraining and task adaptation in StefaLand, we conduct a structured ablation study across multiple tasks and spatial evaluation protocols, comparing four variants against the proposed pretrained model with task adapter.

Direct prediction removes pretraining entirely and treats the target as an additional input channel, training the model end to end only on downstream data.
Scratch training uses the full StefaLand architecture but trains all parameters from random initialization, isolating the effect of large-scale pretraining.
No adapter removes the task-specific adapter and feeds pretrained embeddings directly to the task decoder, testing whether simple feature reuse is sufficient.
Linear probing freezes the pretrained encoder and applies only a linear prediction head, representing the weakest form of adaptation.

Across streamflow, soil moisture, and global Caravan experiments, performance degradation relative to the full model is dominated by the removal of pretraining. Direct prediction and scratch training consistently yield the largest errors, especially under spatial generalization. In contrast, removing or simplifying the adapter produces smaller but consistent performance drops, indicating that task adaptation improves results but cannot compensate for the absence of pretrained representations.

These trends are consistent across random and regional holdouts and across tasks with differing data characteristics, highlighting pretrained land-surface representations as the primary driver of StefaLand’s performance. Numerical results for ablations, are provided in Appendix~\ref{appendix:B:ablation}.

\section{Discussion}
\label{sec:discussion}

\subsection{Key Findings and Contributions}
Methods that reliably improve spatial generalization to data-scarce regions remain rare in the literature \citep{beery2018terra, Gacu2025}. In contrast, StefaLand combined with lightweight task-specific heads achieves state-of-the-art or competitive performance across four broad problem classes: streamflow prediction (both CAMELS and global), soil moisture, soil composition, and landslide susceptibility, while also strengthening the parameterization of differentiable process-based models. Across tasks, the strongest gains arise from architectures that combine pretrained StefaLand representations with explicit temporal modeling via residual connections, indicating that attribute-centric pretraining captures problem-relevant structure while downstream sequence models resolve task-specific dynamics. Together, these results support the premise that foundation models can improve out-of-domain transfer and help democratize prediction quality in data-scarce regions.

Across five dynamic prediction settings, evaluated against benchmark models that reproduce state-of-the-art results reported in prior work, a consistent pattern emerges. Pretraining StefaLand on landscape attributes yields deep representations that are highly relevant to hydrologic and land-surface prediction tasks, enabling stronger spatial generalization than purely supervised approaches such as LSTM-based models or training from scratch. Comparisons with alternative pretrained representations further suggest that problem relevance of the pretraining signal is at least as important as scale alone, particularly for tasks governed by physical and environmental processes rather than visual appearance. We emphasize that this attribute-based approach is complementary to satellite-centric foundation models: while image-based models excel at extracting large-scale visual patterns, StefaLand focuses on structured environmental attributes that are directly aligned with land-surface processes, offering a more targeted and efficient alternative.

A key practical advantage of StefaLand is computational efficiency. The attribute-based pretraining strategy avoids pixel-wise processing and large image volumes, resulting in a compact transformer with roughly 12 million parameters and substantially lower data-management requirements. As a result, StefaLand pretraining can be performed on modest computational budgets, making large-scale spatial generalization experiments feasible without access to specialized infrastructure. While additional adapter and finetuning variants could be explored, the current results already demonstrate strong performance across diverse tasks, and further expansion must balance marginal gains against growing computational and storage costs.


\subsection{Limitations and Future Work}
Several limitations remain. The selection of geological- or ecologically-focused attributes is limited and more can be added to further characterize the subsurface. Two-dimensional (or image-like) data like elevation map can be selectively incorporated using vision transformer heads in the future. 
Expanding the range of targets to include variables such as evapotranspiration, snow water equivalent, and groundwater levels would broaden its applicability. 
Methodologically, advances such as uncertainty-aware prediction heads, and tighter integration with additional process models offer promising avenues to improve calibration and interpretability while preserving efficiency. 
Overall, StefaLand shows that attribute-centric pretraining combined with lightweight temporal or physics heads can deliver strong spatial generalization across geoscientific tasks while remaining computationally accessible. This points toward a practical path for high-quality predictions in regions where they are most needed but data are most limited.

\bibliographystyle{icml2026}
\bibliography{references}
\clearpage
\appendix
\onecolumn

\section{Detailed Model Architecture}
\label{appendix:A}

This appendix provides the complete mathematical formulation of the StefaLand model architecture.

\subsection{Embedding Dynamic and Static Inputs}

StefaLand independently embeds each dynamic and static variable into a latent space. Specifically, for each dynamic variable $c$ at each time step $t$, a two-step nonlinear embedding is applied individually:

\begin{equation}
z_{t,c} = \text{GELU}(x_{t,c}W_{1,c} + b_{1,c})W_{2,c} + b_{2,c}
\end{equation}

where $W_{1,c} \in \mathbb{R}^{1 \times 64}$ and $W_{2,c} \in \mathbb{R}^{64 \times 256}$ are embedding parameters. After embedding all dynamic variables individually, embeddings are stacked and summed across the variable dimension, resulting in a single embedding vector per time step:

\begin{equation}
z_t = \sum_{c=1}^{C} z_{t,c}
\end{equation}

Similarly, static attributes are embedded individually:

\begin{equation}
z_{\text{static},i} = \text{GELU}(s_i W_{1,i} + b_{1,i})W_{2,i} + b_{2,i}
\end{equation}

where separate embedding layers are used for static features. These individual static embeddings are then concatenated with dynamic embeddings along the temporal dimension, resulting in a unified embedding tensor:

\begin{equation}
Z = [z_1; z_2; \ldots; z_T; z_{\text{static}}]
\end{equation}

This static embedding acts as a global learnable token, allowing the model to incorporate basin-specific context into temporal dynamics at any depth of the Transformer layers.

\subsection{Location-aware Cross-Variable Group Masking}

StefaLand introduces Cross-Variable Group Masking (CVGM), a masking strategy that forces the model to capture interactions among correlated hydrologic variables rather than treating them independently. Given a predefinition of hydrological variables into groups $G = \{g_1, g_2, \ldots, g_k\}$, masking occurs as follows:

\begin{enumerate}
\item A temporal masking window $[\tau, \tau + \ell)$ is randomly sampled with length $\ell \sim \mathcal{U}(L_{\text{min}}, L_{\text{max}})$.
\item For each variable group $g_k$, a Bernoulli mask indicator $m_k \sim \text{Bernoulli}(p_{\text{mask}})$ determines whether the group is masked.
\item For each time step $t$ within the masked temporal window and each variable $c$ belonging to masked groups, the embedded feature vector is replaced by a learned mask vector $\mathbf{m}_c$.
\end{enumerate}

Each hydrologic variable $c$ has its own trainable mask embedding vector $\mathbf{m}_c \in \mathbb{R}^{256}$. This CVGM procedure creates reconstruction targets that require modeling cross-variable dependencies and physical interactions.

\subsection{Learnable Positional Encoding}

To provide positional information, StefaLandemploys learnable positional encoding. Each position $i$, corresponding to each time step and the appended static embedding, is assigned a trainable embedding vector $\mathbf{p}_i$. The encoded embedding becomes:

\begin{equation}
\tilde{Z} = Z + P
\end{equation}

where $P = [\mathbf{p}_1; \ldots; \mathbf{p}_{T+1}]$.

\subsection{Transformer Encoder}

The embeddings enriched by positional encoding are processed through an $N$-layer Transformer encoder, where each Transformer block successively applies Multi-Head Self-Attention (MHA) with $h$ attention heads, followed by a residual connection and Layer Normalization. Subsequently, a position-wise Feedforward Network (FFN) is applied, also followed by another residual connection and Layer Normalization:

\begin{align}
A^{(\ell)} &= \text{MHA}(H^{(\ell-1)}) \\
\tilde{H}^{(\ell)} &= \text{LayerNorm}(H^{(\ell-1)} + A^{(\ell)}) \\
F^{(\ell)} &= \text{FFN}(\tilde{H}^{(\ell)}) \\
H^{(\ell)} &= \text{LayerNorm}(\tilde{H}^{(\ell)} + F^{(\ell)})
\end{align}

\subsection{Reconstruction of Original Inputs}

The final hidden states from the Transformer encoder, $H^{(N)}$, are linearly projected and passed through a single-layer bidirectional LSTM to capture the local temporal dependencies and continuity:

\begin{equation}
U = \text{LSTM}(H^{(N)}W_{\text{enc-proj}} + b_{\text{enc-proj}})
\end{equation}

The outputs $U$ are then separated into dynamic and static components, $U_t$ and $U_{\text{static}}$, corresponding to the temporal sequence and static attributes:

\begin{equation}
U_t, U_{\text{static}} = U_{1:T}, U_{T+1}
\end{equation}

Finally, both dynamic and static representations are individually projected back to their original dimensions through separate embedding layers, reconstructing the masked portions of the inputs. Dynamic variables are restored via:

\begin{equation}
\hat{x}_t = \text{DynamicDecEmbedding}(U_t)
\end{equation}

while static attributes are restored by:

\begin{equation}
\hat{s} = \text{StaticDecEmbedding}(U_{\text{static}})
\end{equation}

The projections leverage the learned latent representations to reconstruct the original hydrologic inputs.
\clearpage

\section{Additional Experiments}
\label{appendix:B}

\subsection{Physics-Based Differentiable Modeling}

To leverage domain knowledge and physical constraints inherent in hydrological systems, we implemented physics-based models that explicitly represent hydrological processes through mathematical formulations. These differentiable versions can be trained end-to-end within neural network frameworks, combining process understanding with machine learning flexibility \citep{shen2023differentiable}.

For the process-based backbone, we employed the Hydrologiska Byråns Vattenbalansavdelning (HBV) model \citep{aghakouchak2010,beck2020b,bergstrom1976,bergstrom1992,seibert2012}, a relatively simple bucket-type conceptual hydrologic model. HBV has state variables like snow storage, soil water, and subsurface storage, and can simulate flux variables such as evapotranspiration (ET), recharge, surface runoff, shallow subsurface flow, and groundwater flow. We used an updated modern version, HBV1.1 \citep{song2025physics}, which includes modifications such as increased parallel storage components to represent heterogeneity within basins and dynamic parameterization capabilities.

The hybrid model employs a differentiable parameter learning (dPL) framework where neural networks generate parameters for HBV1.1, and errors are backpropagated through the entire system. A machine learning network takes basin attributes and meteorological forcings as inputs and outputs HBV parameters—both static (e.g., recession coefficients) and dynamic parameters that vary daily. Because HBV1.1 supports automatic differentiation, it serves as the physical backbone: during training, loss is calculated between simulated and observed streamflow, gradients are backpropagated through HBV equations, and neural network weights are updated. This differs from traditional calibration because parameters are learned regionally across all basins simultaneously rather than individually, allowing the network to capture generalizable relationships between basin characteristics and optimal parameters while maintaining mass balance constraints. The system uses 16 parallel response units for spatial heterogeneity and outputs diagnostic variables (e.g., evapotranspiration, soil moisture, baseflow) not directly trained on, providing interpretability with competitive performance.

For physics-based configurations, we tested: (1) a baseline LSTM--HBV1.1 configuration as a standard reference, (2) StefaLand HBV1.1 with resConn, which combines the physics-based approach with our residual connection architecture, and (3) StefaLand HBV1.1 without resConn. These physics-based approaches incorporate hydrological process understanding while maintaining the ability to learn from data..

\begin{table}[!htbp]
\centering
\caption{CAMELS Streamflow PUB and PUR Results (Physics-Based Models)}
\resizebox{\linewidth}{!}{%
\begin{tabular}{l|cccc|cccc}
\toprule
Model & \multicolumn{4}{c|}{Random holdout (ungauged basins)} & \multicolumn{4}{c}{Regional holdout (ungauged regions)} \\
\midrule
 & RMSE $\downarrow$ & µbRMSE $\downarrow$ & Corr $\uparrow$ & NSE $\uparrow$
 & RMSE $\downarrow$ & µbRMSE $\downarrow$ & Corr $\uparrow$ & NSE $\uparrow$ \\
\midrule
LSTM - HBV1.1 & 1.325 & 1.298 & 0.857 & 0.672 & 1.561 & 1.521 & 0.746 & 0.578 \\
StefaLand - resConn HBV1.1 & \textbf{1.234} & \textbf{1.216} & \textbf{0.863} & \textbf{0.714} & \textbf{1.345} & \textbf{1.332} & \textbf{0.842} & \textbf{0.643} \\
StefaLand - no resConn HBV1.1 & 1.315 & 1.302 & 0.848 & 0.707 & 1.401 & 1.379 & 0.835 & 0.623 \\
StefaLand Ablation - resConn HBV1.1 & 1.310 & 1.306 & 0.842 & 0.693 & 1.465 & 1.432 & 0.607 & 0.512 \\
\bottomrule
\end{tabular}%
}
\label{tab:camels_streamflow_physics}
\end{table}

\subsection{Linear Regression baselines}

To justify the use of complex neural networks over traditional methods, we have conducted baseline comparisons using linear regression models. As shown in the table below, linear regression performs poorly across all tasks by a fair margin when compared to our neural network approaches.

\begin{table}[H]
\centering
\caption{Additional experiments with linear regression baselines.}
\label{tab:linear_regression}
\resizebox{\linewidth}{!}{%
\begin{tabular}{cccc|ccc}
\toprule
Experiment & \multicolumn{3}{c|}{Random holdout} & \multicolumn{3}{c}{Regional holdout} \\
\cmidrule(lr){2-4} \cmidrule(lr){5-7}
 & RMSE ↓ & µbRMSE ↓ & Corr ↑ & RMSE ↓ & µbRMSE ↓ & Corr ↑ \\
\midrule
Camels Streamflow Linear Regression & 2.190 & 2.180 & 0.500 & 2.260 & 2.250 & 0.500 \\
Caravan Streamflow Linear Regression & 2.612 & 2.431 & 0.142 & -- & -- & -- \\
Soil Moisture Linear Regression & 0.120 & 0.101 & 0.188 & 0.121 & 0.103 & 0.187 \\
\bottomrule
\end{tabular}%
}
\end{table}

\FloatBarrier

\subsection{External Foundation Model Comparisons}
\label{appendix:B:foundation}

For completeness, we explored several existing foundation models developed for Earth observation and atmospheric applications, including TerraMind, PrithviWxC, and Galileo \citep{jakubik2025terramind,hsu2024geospatial,tseng2025galileo}. All models were evaluated using consistent downstream protocols, with pretrained encoders either frozen or minimally adapted and paired with task-specific heads comparable to those used for StefaLand. These experiments were intended to probe the extent to which representations learned from large-scale EO or atmospheric data transfer to land–surface and hydrologic prediction tasks.

These experiments are not intended as exhaustive benchmarks or as performance upper bounds for the evaluated models, but rather as feasibility probes to understand whether their pretrained representations can be directly repurposed for land–surface and geohazard tasks under lightweight adaptation

Because these models differ substantially in their native input formats and pretraining objectives, task-specific adaptations were required. TerraMind and PrithviWxC were coupled with the same residual adaptation architecture used for StefaLand, with only the added adaptation units trained. Due to the intensive data and storage requirements of PrithviWxC, inputs were restricted to surface-level variables most directly related to land–surface interactions, together with static attributes, while multi-level atmospheric variables were excluded. For context, StefaLand, TerraMind, and PrithviWxC were pretrained on approximately 2, 11, and 27 terabytes of data, respectively.

Galileo was evaluated in the landslide susceptibility setting, where inputs consist of multichannel static environmental attributes rather than multispectral time series. To accommodate this difference, we applied an input adaptation strategy that pooled spatial features along one dimension, projected the 17 environmental channels to 12 channels via a linear layer, and treated the remaining spatial dimension as pseudo-temporal input. We initialized Galileo using its pretrained Transformer encoder (768-dimensional embeddings, four attention layers with 12 heads each, and a feed-forward dimension of 3072), followed by a two-layer classification head (768→256→1) with ReLU activation and dropout (rate 0.3). The resulting model was fine-tuned end-to-end using AdamW (learning rate $10^{-4}$, weight decay 0.01), with a batch size of 128 for 1000 epochs on the same train–test split as other landslide experiments, using gradient clipping with a maximum norm of 1.0.

Collectively, these experiments provide a broad exploratory comparison of how foundation models pretrained on EO imagery or atmospheric data behave when adapted to land–surface and geohazard prediction tasks.

\begin{table}[!htbp]
\centering
\caption{Performance of external foundation model baselines across tasks. All models use frozen pretrained encoders with the same residual adaptation head.}
\resizebox{\linewidth}{!}{%
\begin{tabular}{l|l|ccc}
\toprule
Task & Model & RMSE $\downarrow$ & µbRMSE $\downarrow$ & Corr $\uparrow$ \\
\midrule
CAMELS Streamflow (PUB)
& TerraMind-resConn 
& 1.332 $\pm$ 0.0410 
& 1.301 $\pm$ 0.0375 
& 0.777 $\pm$ 0.0071 \\

CAMELS Streamflow (PUR)
& TerraMind-resConn 
& 1.420 $\pm$ 0.2021 
& 1.398 $\pm$ 0.1932 
& 0.763 $\pm$ 0.0172 \\

\midrule
Soil Moisture (Random)
& TerraMind-resConn 
& 0.083 $\pm$ 0.0021 
& 0.062 $\pm$ 0.0007 
& 0.694 $\pm$ 0.0289 \\
& PrithviWxC-resConn 
& 0.081 $\pm$ 0.0019 
& 0.060 $\pm$ 0.0004 
& 0.703 $\pm$ 0.0390 \\

Soil Moisture (Europe)
& TerraMind-resConn 
& 0.101 
& 0.080 
& 0.519 \\
& PrithviWxC-resConn 
& 0.103 
& 0.079 
& 0.523 \\
\bottomrule
\end{tabular}
}
\label{tab:external_fm}
\end{table}

\begin{table}[!htbp]
\centering
\caption{Landslide susceptibility prediction using the Galileo foundation model compared with published and proposed baselines on the Oregon SLIDO dataset.}
\resizebox{\linewidth}{!}{%
\begin{tabular}{l|ccccc}
\toprule
Model (10m) & Accuracy $\uparrow$ & Precision $\uparrow$ & Recall $\uparrow$ & F1 $\uparrow$ & ROC AUC $\uparrow$ \\
\midrule
Logistic Regression 2D & 0.744 & 0.720 & 0.795 & 0.756 & 0.823 \\
Random Forest 2D       & 0.765 & 0.737 & 0.822 & 0.777 & 0.849 \\
CNN2D                  & 0.880 & \textbf{0.896} & 0.858 & 0.877 & 0.854 \\
Galileo + CNN2D        & 0.750 & 0.764 & 0.720 & 0.742 & 0.834 \\
StefaLand + CNN2D      & \textbf{0.903} & 0.859 & \textbf{0.963} & \textbf{0.908} & \textbf{0.911} \\
\bottomrule
\end{tabular}
}
\label{tab:galileo_landslide}
\end{table}

\subsection{Ablations}
\label{appendix:B:ablation}
\FloatBarrier
\begin{table}[!htbp]
\centering
\caption{StefaLand ablations on CAMELS streamflow under random (PUB) and regional (PUR) spatial holdout.}
\resizebox{\linewidth}{!}{%
\begin{tabular}{l|cccc|cccc}
\toprule
Variant & \multicolumn{4}{c|}{PUB: random holdout (ungauged basins)} & \multicolumn{4}{c}{PUR: regional holdout (ungauged regions)} \\
\midrule
& RMSE $\downarrow$ & ubRMSE $\downarrow$ & Corr $\uparrow$ & NSE $\uparrow$
& RMSE $\downarrow$ & ubRMSE $\downarrow$ & Corr $\uparrow$ & NSE $\uparrow$ \\
\midrule
StefaLand direct      & 1.882 $\pm$ 0.0901 & 1.849 $\pm$ 0.0691 & 0.538 $\pm$ 0.0105 & 0.395 $\pm$ 0.1431
                      & 1.982 $\pm$ 0.4012 & 1.949 $\pm$ 0.3871 & 0.230 $\pm$ 0.0784 & 0.201 $\pm$ 1.2201 \\
StefaLand scratch     & 1.355 $\pm$ 0.0394 & 1.332 $\pm$ 0.0368 & 0.801 $\pm$ 0.0031 & 0.661 $\pm$ 0.0372
                      & 1.516 $\pm$ 0.3723 & 1.378 $\pm$ 0.3496 & 0.771 $\pm$ 0.0211 & 0.560 $\pm$ 0.3120 \\
StefaLand noResConn   & 1.171 $\pm$ 0.0325 & 1.154 $\pm$ 0.0319 & 0.823 $\pm$ 0.0022 & 0.706 $\pm$ 0.3260
                      & 1.376 $\pm$ 0.1987 & 1.356 $\pm$ 0.1712 & 0.798 $\pm$ 0.0195 & 0.610 $\pm$ 0.1345 \\
StefaLand linear      & 1.452 $\pm$ 0.0542 & 1.366 $\pm$ 0.0533 & 0.751 $\pm$ 0.0127 & 0.661 $\pm$ 0.0430
                      & 1.484 $\pm$ 0.2588 & 1.453 $\pm$ 0.2402 & 0.672 $\pm$ 0.0391 & 0.542 $\pm$ 0.3100 \\
StefaLand resConn     & \textbf{1.111 $\pm$ 0.0378} & \textbf{1.068 $\pm$ 0.0374} & \textbf{0.869 $\pm$ 0.0067} & \textbf{0.717 $\pm$ 0.1600}
                      & \textbf{1.344 $\pm$ 0.2097} & \textbf{1.334 $\pm$ 0.1873} & \textbf{0.801 $\pm$ 0.0220} & \textbf{0.635 $\pm$ 0.2460} \\
\bottomrule
\end{tabular}%
}
\label{tab:camels_ablation_pub_pur}
\end{table}

\begin{table}[!htbp]
\centering
\caption{StefaLand ablation study on soil moisture prediction under random and regional holdout.}
\resizebox{\linewidth}{!}{%
\begin{tabular}{l|ccc|ccc}
\toprule
Variant & \multicolumn{3}{c|}{Random location holdout} & \multicolumn{3}{c}{Regional holdout (Europe)} \\
\midrule
 & RMSE $\downarrow$ & ubRMSE $\downarrow$ & Corr $\uparrow$
 & RMSE $\downarrow$ & ubRMSE $\downarrow$ & Corr $\uparrow$ \\
\midrule
StefaLand direct
 & 0.140 $\pm$ 0.0431 & 0.103 $\pm$ 0.0041 & 0.637 $\pm$ 0.0352
 & 0.135 & 0.112 & 0.503 \\
StefaLand scratch
 & 0.074 $\pm$ 0.0011 & 0.058 $\pm$ 0.0003 & 0.749 $\pm$ 0.0172
 & 0.108 & 0.064 & 0.528 \\
StefaLand noResConn
 & 0.075 $\pm$ 0.0009 & 0.057 $\pm$ 0.0001 & 0.741 $\pm$ 0.0201
 & 0.095 & \textbf{0.058} & 0.545 \\
StefaLand linear
 & 0.084 $\pm$ 0.0010 & 0.061 $\pm$ 0.0002 & 0.720 $\pm$ 0.0192
 & 0.100 & 0.063 & 0.393 \\
StefaLand resConn (proposed)
 & \textbf{0.068 $\pm$ 0.0013} & \textbf{0.054 $\pm$ 0.0004} & \textbf{0.783 $\pm$ 0.0054}
 & \textbf{0.090} & 0.059 & \textbf{0.638} \\
\bottomrule
\end{tabular}%
}
\label{tab:soil_moisture_ablation}
\end{table}

\begin{figure}[!htbp]
    \centering
    \includegraphics[width=\textwidth]{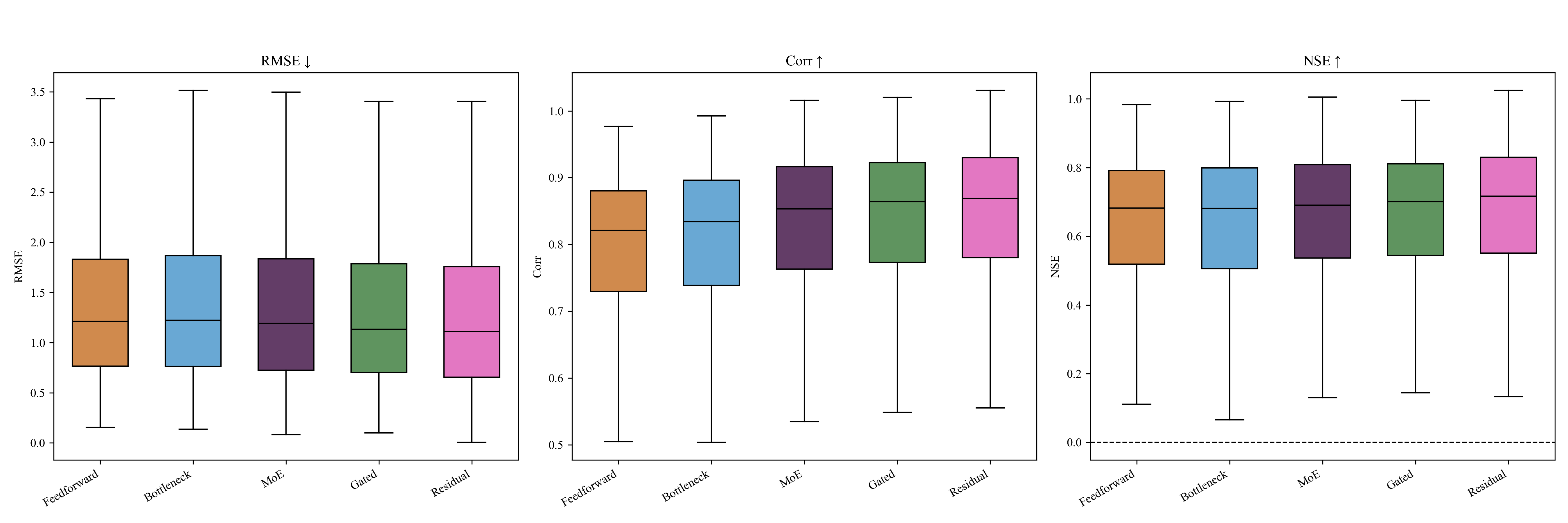}
    \caption{
    Adapter ablation on streamflow random spatial splits (PUB). Boxplots summarize per-basin performance distributions for five adapter designs (Feedforward, Bottleneck, MoE, Gated, Residual). We report RMSE (lower is better), correlation (higher is better), and NSE (higher is better) across held-out basins, showing that the Residual adapter yields the most consistent gains, particularly in Corr and NSE.
    }
    \label{fig:adapter_ablation_pub_boxplots}
\end{figure}

\clearpage

\section{Experimental Details}
\label{appendix:C}


\subsection{Model Configurations and Hyperparameters}
\label{appendix:C:configs}
\begin{table}[!htbp]
\centering
\caption{StefaLand pretraining configuration.}
\small
\setlength{\tabcolsep}{5pt}
\begin{tabularx}{\linewidth}{lX}
\toprule
Parameter & Value \\
\midrule
\multicolumn{2}{l}{\textbf{General Settings}} \\
Task & pretrain \\
Model & Stefaland\_dec\_LSTM \\
Random seed & 111 \\
Time Period & 1980/1/1--2018/12/31 \\
\midrule
\multicolumn{2}{l}{\textbf{Sequence Configuration}} \\
Sequence length & 365 \\
Label length & 365 \\
Prediction length & 365 \\
Sampling stride & 1 \\
Minimum window size & 30 \\
Maximum window size & 90 \\
\midrule
\multicolumn{2}{l}{\textbf{Model Architecture}} \\
Input dimension (enc\_in) & 32 \\
Decoder input (dec\_in) & 6 \\
Output dimension (c\_out) & 6 \\
Model dimension & 256 \\
Number of heads & 4 \\
Encoder layers & 4 \\
Decoder layers & 2 \\
Feed-forward dimension & 512 \\
Dropout & 0.1 \\
Activation & gelu \\
\midrule
\multicolumn{2}{l}{\textbf{Training Configuration}} \\
Optimizer & AdamW \\
Loss criterion & MaskedNSE \\
Epochs & 25 \\
Batch size & 256 \\
Learning rate & 0.0001 \\
Weight decay & 0.0 \\
Patience & 30 \\
Gradient clipping & 5.0 \\
Number of workers & 10 \\
\midrule
\multicolumn{2}{l}{\textbf{Loss Weights}} \\
Time series loss ratio & 1.0 \\
Static loss ratio & 0.5 \\
\bottomrule
\end{tabularx}
\label{tab:pretrain_config}
\end{table}

\begin{table}[!htbp]
\centering
\caption{Attribute groups used in group masking pretraining.}
\small
\setlength{\tabcolsep}{5pt}
\begin{tabularx}{\linewidth}{lX}
\toprule
\textbf{Group} & \textbf{Variables} \\
\midrule
\textbf{Topography} & meanelevation, meanslope \\
\textbf{Soil} & HWSD\_clay, HWSD\_sand, HWSD\_silt, HWSD\_gravel, SoilGrids1km\_sand, SoilGrids1km\_clay, SoilGrids1km\_silt \\
\textbf{Geology} & permeability, Porosity, glaciers, permafrost \\
\textbf{Vegetation} & NDVI, FW \\
\textbf{Climate} & aridity, meanP, ETPOT\_Hargr, meanTa, seasonality\_P, seasonality\_PET, snow\_fraction, snowfall\_fraction \\
\bottomrule
\end{tabularx}
\label{tab:group_masking}
\end{table}

\begin{table}[!htbp]
\centering
\caption{CAMELS streamflow HBV model hyperparameters.}
\small
\setlength{\tabcolsep}{5pt}
\begin{tabularx}{\linewidth}{lX}
\toprule
Parameter & Value \\
\midrule
\multicolumn{2}{l}{\textbf{General Settings}} \\
Random seed & 111111 \\
Data sampler & finetune\_sampler \\
\midrule
\multicolumn{2}{l}{\textbf{Training Configuration}} \\
Time period & 1989/10/01--2008/09/30 \\
Optimizer & Adadelta \\
Batch size & 64 \\
Epochs & 25 \\
\midrule
\multicolumn{2}{l}{\textbf{Neural Model Configuration}} \\
Sequence length & 365 \\
Hidden size & 512 \\
Dropout & 0.2 \\
Encoder layers & 4 \\
Decoder layers & 2 \\
Feed-forward dimension & 512 \\
\midrule
\multicolumn{2}{l}{\textbf{Physical Model (HBV-1.1)}} \\
Model type & HBV\_1\_1p \\
Number of runs (nmul) & 16 \\
Warm-up period & 365 days \\
Warm-up states & True \\
Dynamic dropout & 0.0 \\
Use routing & True \\
Dynamic parameters & parBETA, parK0, parBETAET \\
Near-zero threshold & 1e-05 \\
\midrule
\multicolumn{2}{l}{\textbf{Loss Function}} \\
Type & RmseLoss \\
\bottomrule
\end{tabularx}
\label{tab:camels_hbv_hparams}
\end{table}

\begin{table}[!htbp]
\centering
\caption{Soil moisture model configuration.}
\small
\setlength{\tabcolsep}{5pt}
\begin{tabularx}{\linewidth}{lX}
\toprule
Parameter & Value \\
\midrule
\multicolumn{2}{l}{\textbf{General Settings}} \\
Mode & traintest \\
Random seed & 111111 \\
Data loader & onlylstmloader \\
Data sampler & finetuningnoHBV \\
\midrule
\multicolumn{2}{l}{\textbf{Training Configuration}} \\
Time period & 2015/04/01--2020/12/31 \\
Target & soil\_moisture \\
Optimizer & Adadelta \\
Batch size & 128 \\
Epochs & 50 \\
Save frequency & Every 25 epochs \\
\midrule
\multicolumn{2}{l}{\textbf{Neural Network Configuration}} \\
Hidden size & 128 \\
Dropout & 0.3 \\
Learning rate & 1.2 \\
Encoder layers & 16 \\
Decoder layers & 12 \\
Feed-forward dimension & 512 \\
Rho & 365 \\
\midrule
\multicolumn{2}{l}{\textbf{Loss Function}} \\
Type & RmseLoss \\
\bottomrule
\end{tabularx}
\label{tab:soil_moisture_config}
\end{table}
\FloatBarrier
\subsection{Variables and Data Sources}
\label{appendix:C:variables}


\begin{table}[!htbp]
\centering
\caption{StefaLand pretraining variables and sources.}
\label{table:stefaland_pretraining_vars}
\small
\setlength{\tabcolsep}{6pt}
\begin{tabularx}{\textwidth}{lXX}
\toprule
Variable Type & Variable Name & Source \\
\midrule
\textbf{Time Series Forcings} & Precipitation, Short-wave solar radiation downwards, Relative humidity, Maximum temperature, Minimum temperature, Potential evapotranspiration & from Multi-Source Weather (MSWX) and Multi-Source Weighted-Ensemble Precipitation (MSWEP) \citep{Beck2022MSWX,Beck2019MSWEPv2} \\
\midrule
\textbf{Static Attributes} & Forest cover fraction, grassland cover fraction & Climate Change Initiative (CCI) land cover dataset \citep{esa_land_2017}\\
& Normalized Difference Vegetation Index (NDVI) & Terra Moderate Resolution Imaging Spectroradiometer (MODIS) Vegetation Indices (MOD13A3) \citep{didan_mod13a2_2015}\\
& Sand, silt, clay fractions & Harmonized World Soil Database (HWSD) \citep{fao_harmonized_2012}\\
& Elevation, slope, aspect & Global Multi-resolution Terrain Elevation Data (GMTED) \citep{danielson_global_2011, ramcharan2018}\\
& Soil depth & Global 1-km Gridded Thickness of Soil, Regolith, and Sedimentary Deposit Layers \citep{pelletier_global_2016}\\
& Carbonate sedimentary rock fraction & Global Lithological Map (GLiM) \citep{hartmann_new_2012}\\
& Rock porosity, permeability & GLobal HYdrogeology MaPS (GLHYMPS) \citep{gleeson_glimpse_2014}\\
& Population density & Gridded Population of the World (GPW) v4 dataset \citep{ciesin_gridded_2016}\\
& GDP per capita; population density & Gross Domestic Product and Human Development Index over 1990-2015 \citep{kummu_gridded_2018}\\
& Forest intact fraction & Intact Forest Landscapes Data \citep{potapov_last_2017}\\
\midrule
\textbf{Outputs} & None (self-supervised pretraining) & --- \\
\bottomrule
\end{tabularx}
\end{table}

\begin{table}[!htbp]
\centering
\caption{CAMELS streamflow variables and sources.}
\small
\setlength{\tabcolsep}{6pt}
\begin{tabularx}{\textwidth}{lXX}
\toprule
Variable Type & Variable Name & Source \\
\midrule
\textbf{Time Series Forcings} & Precipitation, Temperature, Potential evapotranspiration, Solar radiation, Vapor pressure & Catchment Attributes and Meteorology for Large-sample Studies (CAMELS) \citep{addor2017camels, newman2014} \\
\midrule
\textbf{Static Attributes} & Elevation, slope, catchment area, forest cover, LAI, GVF, soil depth, porosity, conductivity, sand, silt, clay fractions, carbonate fraction, permeability, aridity, snow fraction, precipitation extremes & CAMELS \\
\midrule
\textbf{Outputs} & Streamflow & CAMELS gauge records \\
\bottomrule
\end{tabularx}
\label{tab:camels_variables}
\end{table}

\begin{table}[!htbp]
\centering
\caption{Soil moisture variables and sources.}
\small
\setlength{\tabcolsep}{6pt}
\begin{tabularx}{\textwidth}{lXX}
\toprule
Variable Type & Variable Name & Source \\
\midrule
\textbf{Time Series Forcings} & Albedo (BSA, WSA) & Moderate Resolution Imaging Spectroradiometer (MODIS) MCD43A3 version 6 \citep{schaaf__crystal_modisterraaqua_2021} \\
& LST (Day, Night) & MODIS Land Surface Temperature/Emissivity Daily (MYD11A1) Version 6.1 \citep{MYD11A1_061} \\
& Precipitation & Global Precipitation Measurement (GPM), MSWEP, and ERA5 precipitation \citep{Huffman2019IMERGv06,Beck2019MSWEPv2,munoz_sabater_era5-land_2019} \\
& Forecast albedo, LAI (high/low vegetation), soil temperature (layer 1), surface pressure, solar radiation, 2\,m temperature, evaporation, precipitation, U/V wind (10\,m) & ECMWF Reanalysis v5 (ERA5) \citep{munoz_sabater_era5-land_2019} \\
\midrule
\textbf{Static Attributes} & elevation, slope, aspect, roughness, curvature & Global 1/5/10/100-km topography derivatives \citep{amatulli_suite_2018} \\
& Sand, clay, silt, bulk density & HWSD v1.2 \citep{fao_harmonized_2012} \\
& Land cover; urban; open water; snow/ice & ESA CCI Land Cover \citep{esa_land_2017} \\
& NDVI & Vegetation Indices Monthly L3 Global 0.05Deg CMG \citep{didan_mod13c2_2015} \\
\midrule
\textbf{Outputs} & Soil moisture & International Soil Moisture Network (ISMN) \citep{dorigo_global_2013, dorigo_international_2011} \\
\bottomrule
\end{tabularx}
\label{tab:soil_moisture_variables}
\end{table}

\begin{table}[!htbp]
\centering
\caption{Streamflow input variables and attributes used from the Caravan dataset.}
\small
\setlength{\tabcolsep}{6pt}
\begin{tabularx}{\textwidth}{lXX}
\toprule
Variable Type & Variable Name & Source \\
\midrule
\textbf{Time Series Forcings} 
& Precipitation (P), Air temperature (Ta), Potential evapotranspiration (PET) 
& ERA5-Land via Caravan \citep{kratzert2023caravan, MunozSabater2021ERA5Land} \\

& Surface pressure, 10 m wind components ($u$, $v$) 
& ERA5-Land via Caravan \citep{kratzert2023caravan, MunozSabater2021ERA5Land} \\

& Net solar radiation, net thermal radiation 
& ERA5-Land via Caravan \citep{kratzert2023caravan, MunozSabater2021ERA5Land} \\

& Snow water equivalent, soil moisture (4 layers) 
& ERA5-Land via Caravan \citep{kratzert2023caravan, MunozSabater2021ERA5Land} \\
\midrule
\textbf{Static Basin Attributes} 
& Latitude, longitude, catchment area 
& Caravan metadata \citep{kratzert2023caravan} \\

& Aridity indices (ERA5-Land, FAO Penman--Monteith) 
& Caravan derived attributes \citep{kratzert2023caravan} \\

& Mean precipitation, precipitation seasonality 
& Caravan derived attributes \citep{kratzert2023caravan} \\

& Mean air temperature, PET seasonality 
& Caravan derived attributes \citep{kratzert2023caravan} \\

& Elevation, slope 
& Global terrain products via Caravan \citep{kratzert2023caravan} \\

& Soil texture fractions (sand, silt, clay) 
& Harmonized World Soil Database (HWSD) via Caravan \citep{fao_harmonized_2012, kratzert2023caravan} \\

& Soil erosion index 
& Global soil datasets via Caravan \citep{kratzert2023caravan} \\

& Forest cover fraction 
& Global land cover products via Caravan \citep{kratzert2023caravan} \\
\midrule
\textbf{Outputs} 
& Streamflow 
& Gauge observations compiled in Caravan \citep{kratzert2023caravan} \\
\bottomrule
\end{tabularx}
\label{tab:caravan_streamflow_variables}
\end{table}

\begin{table}[!htbp]
\centering
\caption{Landslide (SLIDO, Oregon) variables and sources.}
\small
\setlength{\tabcolsep}{6pt}
\begin{tabularx}{\textwidth}{lXX}
\toprule
Variable Type & Variable Name & Source \\
\midrule
\textbf{Input data} & Elevation & National Elevation Dataset (NED) \citep{american_society_for_photogrammetry_and_remote_sensing_national_2018} \\
& Soil sand, silt, clay, bulk density, saturated hydraulic conductivity & Probabilistic Remapping of SSURGO (POLARIS) \citep{chaney_polaris_2019} \\
& Lithology & Global Lithological Map (GLiM) \citep{hartmann_new_2012} \\
& Rainfall & PRISM \citep{prism_climate_group_prism_2014} \\
& NDVI & Moderate Resolution Imaging Spectroradiometer (MODIS) Vegetation Indices Monthly L3 \citep{didan_mod13q1_2015} \\
& Landcover & National Land Cover Database (NLCD) 2016 \citep{dewitz_national_2019} \\
& Soil moisture & SMAP-HydroBlocks (SMAP-HB) \citep{Vergopolan2021SMAPHB} \\
& slope, aspect, curvature, TWI, SPI & DEM-derived \\
\midrule
\textbf{Outputs} & Landslide occurrence (binary) & Statewide Landslide Information Database for Oregon (SLIDO) \citep{franczyk_j_j_statewide_2020} \\
\bottomrule
\end{tabularx}
\label{tab:landslide_variables}
\end{table}

\begin{table}[!htbp]
\centering
\caption{Soil composition (ISRIC) variables and sources.}
\small
\setlength{\tabcolsep}{6pt}
\begin{tabularx}{\textwidth}{lXX}
\toprule
Variable Type & Variable Name & Source \\
\midrule
\textbf{Time Series Forcings} & Same as Table~\ref{tab:soil_moisture_variables} & --- \\
\midrule
\textbf{Static Attributes} & Same as Table~\ref{tab:soil_moisture_variables} & --- \\
\midrule
\textbf{Outputs} & Soil property (clay; sand; silt) & World Soil Information Service (WoSIS) \citep{Batjes2020WoSISsnapshot2019} \\
\bottomrule
\end{tabularx}
\label{tab:soil_composition_variables}
\end{table}

\clearpage
\FloatBarrier
\subsection{Computational Resources}
\label{appendix:C:compute}

\begin{table}[!htbp]
\centering
\caption{Computation Resources for StefaLand and Comparison Models}
\small
\setlength{\tabcolsep}{6pt}
\begin{threeparttable}
\begin{tabularx}{\textwidth}{l c c c c}
\toprule
\textbf{Model} & \textbf{Seconds/Epoch} & \textbf{\#GPUs} & \textbf{GPU Type} & \textbf{Memory} \\
\midrule
StefaLand (Pretraining) & 16{,}000 & 6 & NVIDIA V100 & 240 GB \\
StefaLand ResConn & 30 & 2 & NVIDIA V100 & 80 GB \\
StefaLand no Adapter & 26 & 2 & NVIDIA V100 & 80 GB \\
LSTM Baseline & 12 & 2 & NVIDIA V100 & 80 GB \\
LSTM-HBV1.1 & 280 & 2 & NVIDIA V100 & 80 GB \\
StefaLand-resConn HBV1.1 & 320 & 2 & NVIDIA V100 & 80 GB \\
StefaLand-no Adapter HBV1.1 & 300 & 2 & NVIDIA V100 & 80 GB \\
\bottomrule
\end{tabularx}
\begin{tablenotes}[flushleft]
\small
\item Note: All values except pretraining are for the CAMELS benchmark experiment. Relative computational differences are consistent across other experiments.
\end{tablenotes}
\end{threeparttable}
\label{tab:computation-resources}
\end{table}
\FloatBarrier


\subsection{Pretraining Data Handling}

\label{appendix:pretrain_handling}
\begin{figure*}[!t]
    \centering
    \includegraphics[width=\textwidth]{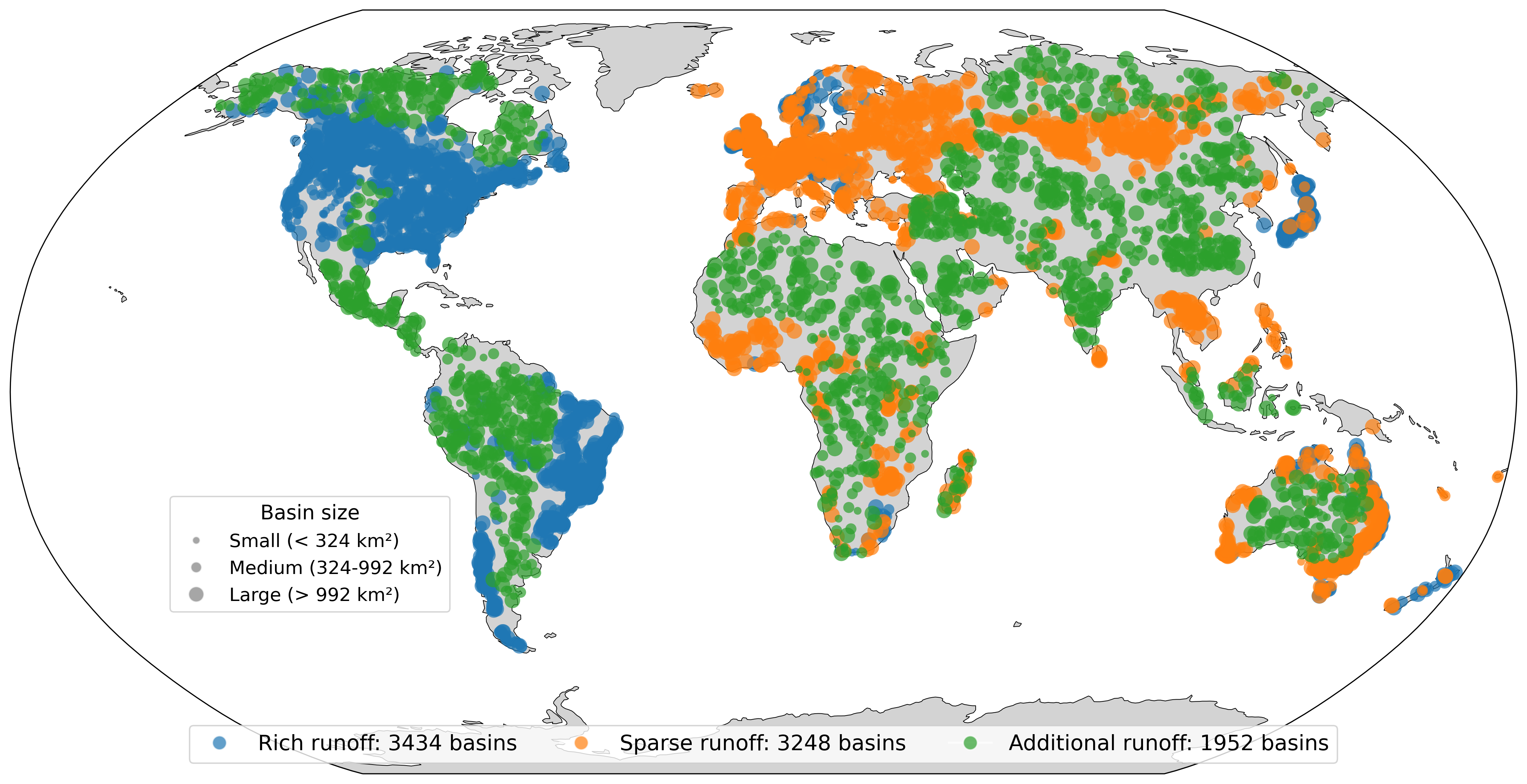}
    \caption{
    Spatial distribution of the global streamflow dataset. Basins are categorized according to the availability of runoff observations: basins with relatively abundant runoff records (blue), basins with sparse runoff records (orange), and basins without runoff data (green). Marker size corresponds to basin area, classified into three categories based on the 33rd and 67th percentiles of catchment areas.
    }
    \label{fig:GLOBAL_data_distribution}
\end{figure*}
A global dataset was constructed for model pretraining, including 8,634 catchments and designed to characterize climatic, ecological, soil, topographic, geological, and socioeconomic conditions. The dataset includes both daily meteorological forcings and long-term averaged static attributes. Daily meteorological variables comprise precipitation, downward shortwave radiation, relative humidity, maximum temperature, and minimum temperature, derived from the Multi-Source Weather (MSWX) and Multi-Source Weighted-Ensemble Precipitation (MSWEP) datasets at a spatial resolution of 0.1° \citep{Beck2022MSWX,Beck2019MSWEPv2}. Potential evapotranspiration was estimated using the Hargreaves method.

The study catchments are divided into three groups: 3,434 GRDC catchments with relatively abundant historical runoff records, 3,248 GRDC catchments with sparse runoff records, and 1,952 HydroBasins level-8 catchments without runoff observations \citep{grdc_global_2024, lehner2013global}. Ecosystem states are represented by forest and grassland cover fractions derived from the Climate Change Initiative (CCI) land cover dataset \citep{esa_land_2017}, along with the Normalized Difference Vegetation Index (NDVI) from MODIS \citep{didan_mod13a2_2015}. Soil properties include sand, silt, and clay fractions from the Harmonized World Soil Database (HWSD) \citep{fao_harmonized_2012}.

Topographic attributes include elevation, slope, and aspect obtained from the Global Multi-resolution Terrain Elevation Data (GMTED) \citep{danielson_global_2011, ramcharan2018}, as well as terrain-derived soil depth from the Global 1-km Gridded Thickness of Soil, Regolith, and Sedimentary Deposit Layers dataset \citep{pelletier_global_2016}. Geological attributes comprise carbonate sedimentary rock fractions from the Global Lithological Map (GLiM) \citep{hartmann_new_2012} and rock porosity and permeability from the GLobal HYdrogeology MaPS (GLHYMPS) dataset \citep{gleeson_glimpse_2014}.

Socioeconomic conditions are characterized using population density from the Gridded Population of the World (GPW) v4 dataset \cite{ciesin_gridded_2016}, gross domestic product and population data from the gridded global GDP and Human Development Index datasets \citep{kummu_gridded_2018}, and forest intactness from the Intact Forest Landscapes dataset \citep{potapov_last_2017}. All static attributes were mapped to a common 0.01° grid prior to basin-scale aggregation to ensure spatial consistency across data sources and improve spatial averaging over irregular basin geometries. A complete list of variables is provided in Table~\ref{table:stefaland_pretraining_vars}.

\subsection{Dataset Splitting}
\label{appendix:dataset_splitting}

For the WoSIS soil dataset, we collected soil property data from 106,503 locations. After removing low-quality records (e.g., sand values greater than 1 or negative values), we randomly sampled 5,000 soil points to reduce computational cost. We then applied 5-fold cross-validation (k=5) on this subset.

For the landslide dataset at 30 m resolution, we used 14,604 historical landslide points. We split the dataset into 70\% for training, 20\% for validation, and 10\% for testing.

\clearpage

\section{Additional Figures}
\label{appendix:D}

\begin{figure}[!htbp]
    \centering
    \includegraphics[width=0.9\textwidth]{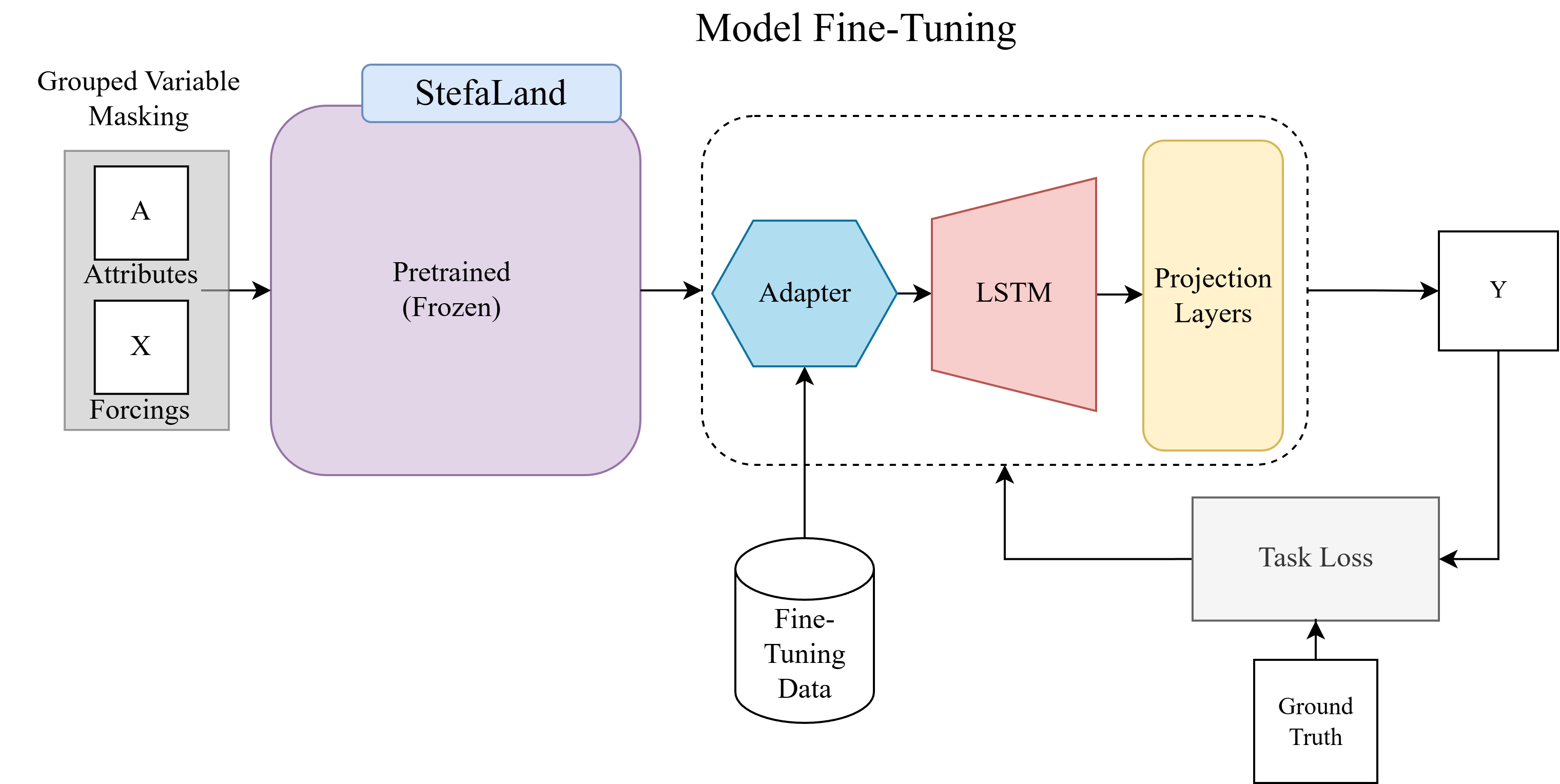}
    \caption{Fine-tuning pipeline used in our downstream experiments. Static attributes and meteorological forcings are encoded by the pretrained StefaLand encoder (frozen), then passed through a task adapter and sequence model (LSTM), followed by projection layers to generate predictions optimized with a task loss against ground truth.}
    \label{fig:StefaLand_finetune}
\end{figure}

\begin{figure}[!htbp]
    \centering
    \includegraphics[width=0.9\textwidth]{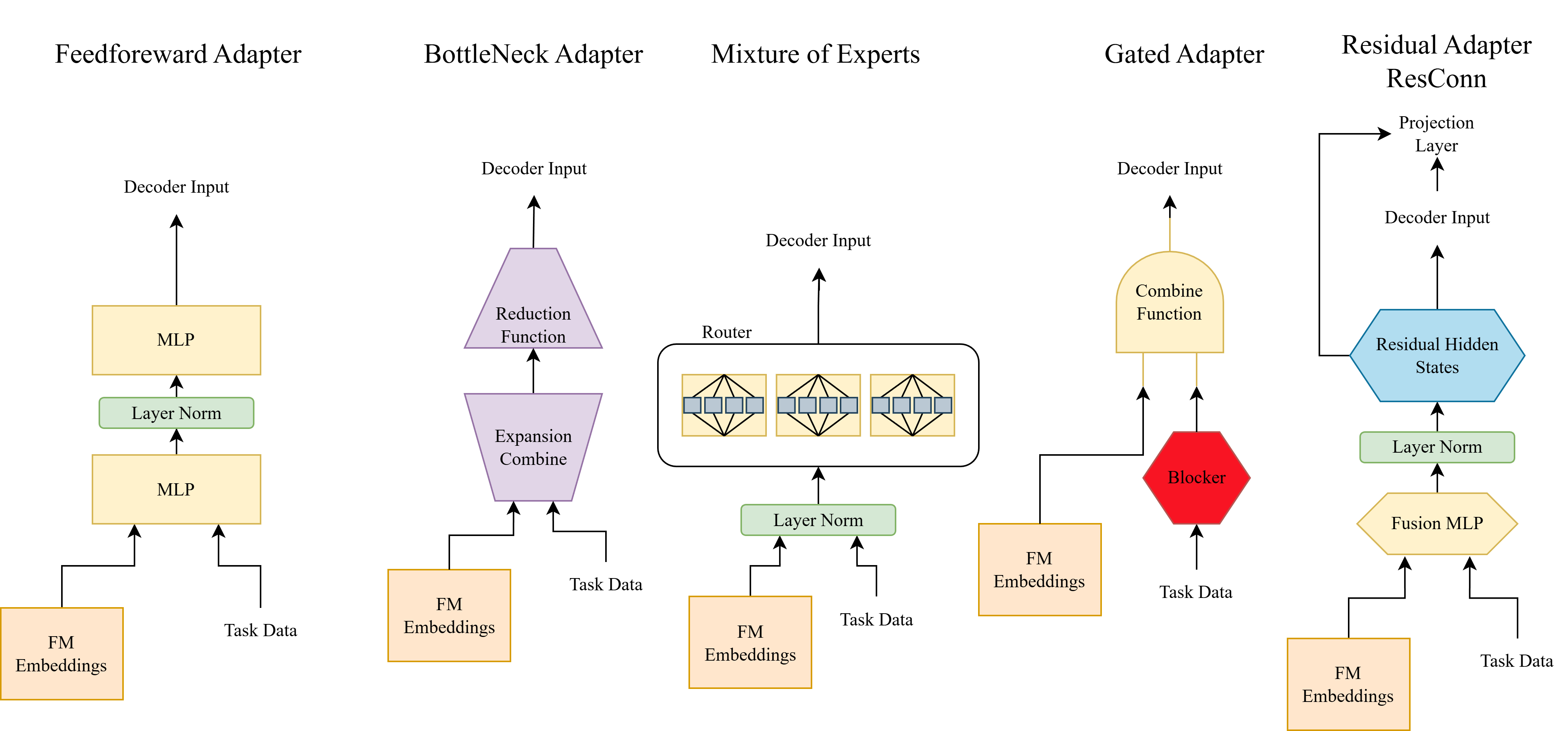}
    \caption{Adapter architectures evaluated in our experiments. We compare a gated adapter, a bottleneck adapter with compression and expansion stages, A mixture of Experts and a basic feedforeward  and a residual adapter that injects pretrained features via a skip connection.}
    \label{fig:adapters}
\end{figure}

\clearpage

\section{Metric Calculations}
\label{appendix:E}

This appendix details the calculation of the evaluation metrics used in our experiments. All metrics presented in the main paper tables are the median values across test basins or stations, as computed using the following formulations.

\subsection{Primary Evaluation Metrics}

\subsubsection{Root Mean Square Error (RMSE)}
RMSE measures the average magnitude of prediction errors. Lower values indicate better performance.

\begin{equation}
\text{RMSE} = \sqrt{\frac{1}{n}\sum_{i=1}^{n}(y_{\text{pred},i} - y_{\text{target},i})^2}
\end{equation}

\subsubsection{Unbiased Root Mean Square Error (µbRMSE)}
µbRMSE removes the bias component from the error calculation, focusing on the error's random component. It is calculated by first computing anomalies from the mean for both predictions and targets.

\begin{align}
y'_{\text{pred},i} &= y_{\text{pred},i} - \overline{y}_{\text{pred}} \\
y'_{\text{target},i} &= y_{\text{target},i} - \overline{y}_{\text{target}} \\
\text{µbRMSE} &= \sqrt{\frac{1}{n}\sum_{i=1}^{n}(y'_{\text{pred},i} - y'_{\text{target},i})^2}
\end{align}

\subsubsection{Correlation (Corr)}
Correlation quantifies the linear relationship between predictions and targets. Values range from -1 to 1, with 1 indicating perfect positive correlation.

\begin{equation}
\text{Corr} = \frac{\sum_{i=1}^{n}(y_{\text{pred},i} - \overline{y}_{\text{pred}})(y_{\text{target},i} - \overline{y}_{\text{target}})}{\sqrt{\sum_{i=1}^{n}(y_{\text{pred},i} - \overline{y}_{\text{pred}})^2 \sum_{i=1}^{n}(y_{\text{target},i} - \overline{y}_{\text{target}})^2}}
\end{equation}

This is calculated using Pearson's correlation coefficient between predicted and observed values.

\subsection{Secondary Metrics}

The following metrics are used in our comprehensive evaluation but may not appear directly in the main tables.

\subsubsection{Nash-Sutcliffe Efficiency (NSE) / $R^2$}
NSE evaluates the predictive skill relative to using the mean of observations as a predictor. Values range from $-\infty$ to 1, with 1 indicating perfect prediction.

\begin{equation}
\text{NSE} = 1 - \frac{\sum_{i=1}^{n}(y_{\text{target},i} - y_{\text{pred},i})^2}{\sum_{i=1}^{n}(y_{\text{target},i} - \overline{y}_{\text{target}})^2}
\end{equation}

\subsubsection{Mean Absolute Error (MAE)}
MAE measures the average absolute difference between predictions and targets.

\begin{equation}
\text{MAE} = \frac{1}{n}\sum_{i=1}^{n}|y_{\text{pred},i} - y_{\text{target},i}|
\end{equation}

\subsubsection{Flow Duration Curve RMSE (RMSE\_FDC)}
RMSE\_FDC evaluates errors in the statistical distribution of flows rather than in their timing.

\begin{equation}
\text{RMSE\_FDC} = \sqrt{\frac{1}{100}\sum_{j=1}^{100}(FDC_{\text{pred},j} - FDC_{\text{target},j})^2}
\end{equation}

where $FDC_j$ represents the $j$-th percentile of the sorted flow values.

\subsubsection{Flow Biases}
Several flow-specific biases were computed to evaluate performance across different flow regimes:

\begin{itemize}
  \item $FLV$ (Low Flow Volume Bias): Percent bias in the lowest 30\% of flows
  \item $FHV$ (High Flow Volume Bias): Percent bias in the highest 2\% of flows
  \item $PBIAS$ (Percent Bias): Overall percent bias across all flows
\end{itemize}

The general form for these biases is:
\begin{equation}
\text{PBIAS}_{\text{regime}} = \frac{\sum(y_{\text{pred},\text{regime}} - y_{\text{target},\text{regime}})}{\sum y_{\text{target},\text{regime}}} \times 100\%
\end{equation}

\subsubsection{Kling-Gupta Efficiency (KGE)}
KGE combines correlation, bias, and variability components:

\begin{equation}
\text{KGE} = 1 - \sqrt{(r - 1)^2 + \left(\frac{\sigma_{\text{pred}}}{\sigma_{\text{target}}} - 1\right)^2 + \left(\frac{\mu_{\text{pred}}}{\mu_{\text{target}}} - 1\right)^2}
\end{equation}

where $r$ is the correlation coefficient, $\sigma$ represents standard deviation, and $\mu$ represents the mean.

\subsection{Metric Aggregation}
For each evaluation scenario (Random Holdout and Regional Holdout), metrics were calculated for each individual basin or station and then aggregated using median values to provide a robust measure of central tendency less sensitive to outliers. All metrics shown in tables throughout the paper represent these median values across the test set.

\subsection{Implementation Details}
All metrics were implemented in Python using NumPy for numerical computations and SciPy's statistical functions for correlation coefficients. Special care was taken to handle missing values (NaNs) appropriately in all calculations. For time series with missing values, only timestamps where both predicted and target values were available were used in metric calculations.

\end{document}